\documentclass{article}
\PassOptionsToPackage{dvipsnames,table}{xcolor}
\usepackage{iclr2027_conference,times}

\usepackage{amsmath,amsfonts,bm}

\def\eqref#1{equation~\ref{#1}}

\def\1{\bm{1}}

\DeclareMathAlphabet{\mathsfit}{\encodingdefault}{\sfdefault}{m}{sl}
\SetMathAlphabet{\mathsfit}{bold}{\encodingdefault}{\sfdefault}{bx}{n}

\usepackage[utf8]{inputenc}
\usepackage[T1]{fontenc}
\usepackage{url}
\usepackage{booktabs}
\usepackage{amsfonts}
\usepackage{nicefrac}
\usepackage{microtype}
\usepackage[dvipsnames, table]{xcolor}
\usepackage{wrapfig}
\usepackage{graphicx}
\usepackage{amsmath,amssymb}
\usepackage{multirow}
\usepackage{tikz}
\usetikzlibrary{arrows.meta,backgrounds}
\usetikzlibrary{tikzmark,calc}
\definecolor{oursorange}{HTML}{FA500E}
\newcommand{\oursL}[1]{\tikzmark{#1-l}}
\newcommand{\oursR}[1]{\tikzmark{#1-r}}
\newcommand{\oursB}[1]{\tikzmark{#1-b}}
\newcommand{\oursdraw}[4]{%
  \begin{tikzpicture}[remember picture, overlay]
    \coordinate (oursl) at (pic cs:#4-l);
    \coordinate (oursr) at (pic cs:#4-r);
    \coordinate (oursb) at (pic cs:#4-b);
    \fill[oursorange!10, rounded corners=3pt]
      ($(oursl) + (-#3, #1)$) rectangle ($(oursr |- oursb) + (#3, -#2)$);
  \end{tikzpicture}}
\NewDocumentCommand{\oursbox}{O{8.5pt} O{4.2pt} O{3pt} m}{%
  \ifvmode
    \par\edef\ourssavedprevdepth{\the\prevdepth}\nointerlineskip
    \hbox{\oursdraw{#1}{#2}{#3}{#4}}%
    \prevdepth=\ourssavedprevdepth\relax
  \else
    \oursdraw{#1}{#2}{#3}{#4}%
  \fi}

\usepackage{subcaption}
\usepackage{pifont}

\usepackage{soul}

\newcommand{\cmark}{\ding{51}}
\newcommand{\xmark}{\ding{55}}

\definecolor{accentorange}{HTML}{FA500E}
\definecolor{accentyellow}{HTML}{FFAE00}
\definecolor{accentblue}{HTML}{10069F}
\definecolor{accentlight}{HTML}{E6F3FB}
\definecolor{burgundy}{rgb}{0.5, 0.0, 0.13}
\definecolor{blueish}{RGB}{176,224,230}
\definecolor{C4}{RGB}{0,148,181}
\definecolor{B4}{RGB}{14,135,201}
\definecolor{myblue}{HTML}{0064E0}
\newcommand\myshade{100}
\colorlet{mylinkcolor}{B4}
\colorlet{myurlcolor}{C4}
\definecolor{cblue}{RGB}{70,130,210}
\definecolor{cgreen}{RGB}{60,160,90}
\definecolor{cpurple}{RGB}{140,90,190}

\usepackage[
    citecolor=accentorange!\myshade!black,
    colorlinks=true,
    linkcolor=accentblue!\myshade!black
]{hyperref}
\newcommand{\mistral}{\raisebox{-0.1\height}{\includegraphics[scale=0.065,  clip]{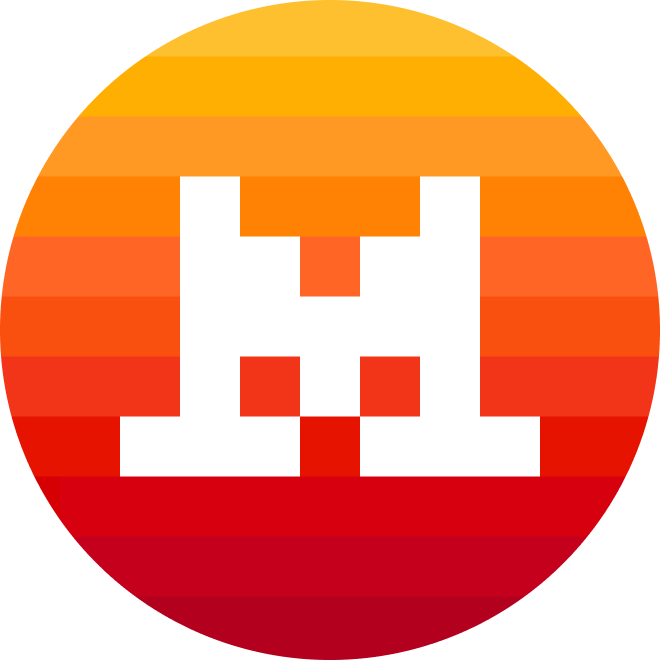}}}

\title{One-Way Coupling via Patchwise Adaptive Normalization in Transformer Surrogates}

\author{
\parbox{\linewidth}{
  Emmanuel Gottlob\textsuperscript{$\dagger$,*} \quad  Hugo Thimonier\textsuperscript{$\dagger$} \hspace{.8cm} Ludovico Buizza \\
  August Martens \quad {\hspace{.8cm}Alexandre Gavaudan}
} \vspace{0.1cm} \\
\hspace{.02cm} Mistral AI \href{https://mistral.ai/}{\mistral} \\
\texttt{\{name\}.\{surname\}@mistral.ai } \\
\And
Thiago Guzella \vspace{0.1cm} \\
ASML The Netherlands B.V. \\
\texttt{thiago.guzella@asml.com} \\
\\
\textsuperscript{$\dagger$} Equal contribution.
\textsuperscript{*} Corresponding author.
}

\preprint
\begin{document}

\maketitle

\begin{abstract}
Many physical systems exhibit \emph{one-way coupling}: exogenous fields such as medium density or obstacle geometry influence the dynamics without being affected by them. We investigate how explicitly representing these fields can improve the accuracy and computational efficiency of transformer-based PDE surrogates. Building on spatially adaptive normalization, we introduce \textbf{Patchwise Adaptive Normalization} (\texttt{PAdaNorm}), which partitions exogenous fields into patches aligned with the state tokenization.
We evaluate \texttt{PAdaNorm} on \textit{datasets from three different domains}, using a Walrus-inspired transformer architecture.
Across the three PDE families, \texttt{PAdaNorm} is on par with or up to 16\% better than parameter-matched baselines in mean long-horizon variance-scaled RMSE on 60-step autoregressive rollouts. We also find that \texttt{PAdaNorm} better maintains hard physical constraints over long rollouts, such as preventing flow into impenetrable solid regions in porous media.
When the conditioning is static, \texttt{PAdaNorm} caches the conditioning embeddings once and reuses them across the entire rollout, reducing inference FLOPs per step by up to 36\% compared with parameter-matched baselines.
An ablation over eight conditioning variants shows that modulating a single attention axis (temporal or spatial) suffices and that a lightweight shared MLP across blocks remains competitive with per-block projections.
\end{abstract}

\section{Introduction}
\label{sec:introduction}

    Transformer-based surrogates have shown promise for learning the evolution of physical fields across diverse systems governed by partial differential equations (PDEs)~\citep{mccabe2024multiplephysicspretrainingphysical,herde2024poseidon,mccabe2025walruscrossdomainfoundationmodel}. Maintaining prediction accuracy over long autoregressive rollouts remains challenging, as errors introduced at individual steps can propagate and accumulate~\citep{brandstetter2022message,lippe2023pderefiner,koehler2024apebench}. One contributing factor is that the model is trained on ground-truth input frames but, during a rollout, receives its own earlier predictions as inputs, whose distribution slowly shifts away from the one seen in training~\citep{bengio2015scheduled,sanchezgonzalez2020learning}. Existing approaches mitigate this rollout degradation through input noise during training~\citep{sanchezgonzalez2020learning,pfaff2021learning}, training over multiple prediction steps~\citep{brandstetter2022message,koehler2024apebench}, or patch jittering~\citep{mccabe2025walruscrossdomainfoundationmodel}. Alongside these approaches, we investigate how conditioning on known spatial inputs, such as material properties and obstacle geometry, affects autoregressive prediction accuracy.

    In many PDE models, the evolving state depends on spatial inputs such as material properties, obstacle geometry, or externally imposed fields. We refer to these inputs as \emph{exogenous fields} when they influence the predicted dynamics without being affected by them, giving rise to \emph{one-way coupling}. For example: (i) thermal stress in a material where a temperature distribution causes the material to expand and develop internal mechanical stresses, while mechanical stress does not impact the temperature profile; or (ii) quantum physics, whereby an external potential influences the dynamics of a particle; or (iii) in photonics, where the refractive index of the material bends the path of light rays.

    Existing PDE surrogates incorporate information about the physical system through several complementary mechanisms. Models such as $\text{GP}_{\text{hy}}\text{T}$ use histories of physical states as context from which to infer the underlying dynamics~\citep{wiesner2026physicsfoundationmodel}. When conditioning fields are explicitly available, they can be concatenated with the state channels, preserving their spatial correspondence at the input. Other architectures introduce separate conditioning pathways: GNOT uses heterogeneous cross-attention to incorporate multiple input functions~\citep{hao2023gnotgeneralneuraloperator}, while GINOT encodes geometry and integrates it with query-point representations through cross-attention~\citep{liu2025geometryinformedneuraloperatortransformer}. Adaptive normalization provides another mechanism for injecting conditioning into intermediate features: instead of learning fixed scale and shift parameters, a normalization layer predicts them from a conditioning signal, as in the adaptive layer normalization (AdaLN) of the Diffusion Transformer~\citep{peebles2023dit}. Poseidon uses lead-time-conditioned layer normalization~\citep{herde2024poseidon}, whereas UPT employs DiT-style scale, shift, and gate modulation to condition on time and boundary-condition parameters such as inflow velocity~\citep{alkin2024upt,peebles2023dit}.

    Spatially varying modulation has precedents in both image synthesis and physical modeling. SPADE generates spatial scale and shift parameters from semantic segmentation masks to modulate convolutional feature maps~\citep{park2019spade}. FlowAR's Spatial-AdaLN produces position-specific scale, shift, and gate parameters from aligned semantic representations within a transformer~\citep{ren2025flowar}. P3D uses region tokens derived from globally aggregated state features to modulate local decoder regions through adaptive normalization~\citep{holzschuh2025p3d}. These methods establish spatial modulation as a mechanism for structured conditioning. Building on this principle, we investigate a specific design for autoregressive PDE surrogates: independently encoding exogenous field patches aligned with the state tokenization, sharing the nonlinear encoder across blocks, and using block-specific projections to modulate the spatial attention and feed-forward block using spatially-aware conditioning.

    We call this design \textbf{Patchwise Adaptive Normalization} (\texttt{PAdaNorm}; Fig.~\ref{fig:padaln}), a conditioning approach for one-way coupled fields in physical PDEs.
    \texttt{PAdaNorm} conditions the transformer on spatially-varying material properties (e.g., density, refractive index, potential strength, ...) at the granularity of individual patches. The conditioning fields are partitioned into the same non-overlapping patches as the input state, and each patch is independently mapped through a shared MLP to produce a conditioning vector. This yields one conditioning vector per spatial patch, capturing the local effect of the coupled exogenous variables within each patch's receptive field. These patchwise conditioning vectors then drive an adaLN-Zero-style shift, scale, and gate modulation~\citep{peebles2023dit} of the transformer's intermediate representations on a per-patch basis: each patch token receives its own shift, scale, and gate parameters derived from the conditioning fields at that spatial location. For static conditioning fields and fixed patch alignment, all modulation parameters can be computed once and reused throughout, amortizing parameter-generation cost over the rollout.

    We evaluate \texttt{PAdaNorm} on several PDE modeling tasks, spanning different PDE families (acoustic scattering, 2D Navier-Stokes, and CO\textsubscript{2}--water flow in porous media). Our experiments demonstrate that \texttt{PAdaNorm} outperforms or closely matches existing baselines, with a comparable parameter count and with fewer forward-pass FLOPs than parameter-matched baselines (App.~\ref{app:flops}). We also find that \texttt{PAdaNorm} better maintains hard physical constraints over long rollouts: on CO\textsubscript{2}--water flow in porous media, it lets significantly less fluid into the impenetrable rock than every baseline we tested (Sec.~\ref{sec:experiments}). Crucially, when the conditioning variables are static, each rollout step of \texttt{PAdaNorm} costs the same matrix-multiplication FLOPs as the same transformer without conditioning; only the element-wise modulation remains.

\section{Related Works}
\label{sec:related_works}

    \paragraph{Neural surrogates for PDEs}
        Neural operators learn mappings between physical fields, with the Fourier Neural Operator parameterizing integral kernels in Fourier space to model families of PDE solutions~\citep{li2021fourierneuraloperator}. Recent work has expanded toward pretraining across heterogeneous physical systems. MPP learns transferable representations through a shared transformer backbone~\citep{mccabe2024multiplephysicspretrainingphysical}, Poseidon uses a multiscale operator transformer~\citep{herde2024poseidon}, and DPOT introduces autoregressive denoising pretraining~\citep{hao2024dpot}. Walrus further develops this direction by introducing a transformer that supports diverse physical fields in two and three spatial dimensions~\citep{mccabe2025walruscrossdomainfoundationmodel}. Our work investigates conditioning mechanisms within a Walrus-inspired backbone, focusing on how spatially varying exogenous fields can condition its intermediate representations.

    \paragraph{Conditioning mechanisms}
        Feature-wise Linear Modulation~\citep{perez2018film} formalizes the general idea of conditioning neural networks by predicting per-feature scale and shift parameters from an auxiliary signal, connecting a family of prior adaptive normalization methods~\citep{dumoulin2018featurewise}.
        In diffusion models, \citet{dhariwal2021diffusionbeatgans} use Adaptive Group Normalization (AdaGN) to condition U-Net layers on timestep and class embeddings.
        The Diffusion Transformer~\citep{peebles2023dit} adapts this mechanism for transformers with AdaLN-Zero, modulating the LayerNorm parameters based on a conditioning vector and initializing each residual block as the identity through zero-initialized gating. Crucially, standard AdaLN applies uniform modulation to all tokens~\citep{chen2024gentron}, which suffices for global signals such as timesteps or class labels but creates a representational bottleneck when the conditioning varies spatially. In image generation, SPADE~\citep{park2019spade} predicts spatially varying scale and shift from semantic segmentation masks to modulate convolutional features, and FlowAR~\citep{ren2025flowar} and PixelDiT~\citep{yu2025pixeldit} bring such spatially adaptive modulation to transformers. \texttt{PAdaNorm} builds on these mechanisms by independently encoding exogenous field patches aligned with the state tokenization. Its nonlinear patch encoder is shared across transformer blocks, while block-specific linear projections generate the modulation parameters. For static fields and fixed patch alignment, these parameters can be cached throughout an autoregressive rollout.

    \paragraph{Conditioning mechanisms for PDE surrogates}
        The simplest approach, used by Walrus~\citep{mccabe2025walruscrossdomainfoundationmodel} and MPP~\citep{mccabe2024multiplephysicspretrainingphysical}, treats conditioning variables as additional input channels concatenated to the state tokens (\emph{in-context conditioning}), without any architectural inductive bias to disentangle state from forcing variables.
        Cross-attention is an alternative, allowing the system's state tokens to attend over the conditioning fields, at a quadratic cost in the number of patches (App.~\ref{app:flops}) and without enforcing spatial or temporal locality.
        Poseidon~\citep{herde2024poseidon} conditions on continuous time through time-dependent LayerNorm~\citep{ba2016layernormalization} parameters, enabling evaluation at arbitrary temporal resolution.
        PDE-Transformer~\citep{holzschuh2025pdetransformer} uses AdaLN-Zero to modulate transformer blocks based on a learned PDE-type embedding, while UPT conditions on time and boundary-condition parameters~\citep{alkin2024upt}.
        In recent work, \citet{holzschuh2025p3d} propose P3D, which applies adaptive instance normalization to local decoder regions using region tokens derived from globally aggregated state features.
        \texttt{PAdaNorm} instead derives each patch's modulation directly from the corresponding exogenous field patch, independently of the evolving state. This imposes locality on the conditioning pathway while allowing the transformer's attention to model interactions between spatial locations.

\section{Method}
\label{sec:method}

    \subsection{Walrus Transformer}
    \label{subsec:walrus}
        Building on its success as a foundation physics model, we rely on a minimal version of Walrus~\citep{mccabe2025walruscrossdomainfoundationmodel}, retaining its core components to minimize confounding factors. We depart from the original implementation in three ways. First, we include neither patch jitter nor adaptive-compute patching. Second, temporal position is encoded with a 1D rotary embedding rather than the learned T5-style relative position encoding. Third, temporal attention carries no causal mask, which was originally found to have no impact on performance compared to causal masking ~\citep[App.~B.3]{mccabe2025walruscrossdomainfoundationmodel}.

        Our implementation of Walrus differs from the vanilla transformer as it includes successive (i) temporal attention and (ii) spatial attention blocks at each layer. See App.~\ref{app:subsec:walrus_transformer} for details on spatial and temporal attention and \citet{mccabe2025walruscrossdomainfoundationmodel} for a full description of the Walrus architecture. Formally, the model predicts the evolution of a 2D state $\mathbf{u}(x,y;\mathbf{C};t+\Delta t)$ at a fixed timestep $\Delta t$ and subject to the spatially varying, static one-way coupled fields $\mathbf{C}(x,y)$ (of dimension $C_{\mathrm{cond}}$) given $T_{\mathrm{in}}$ consecutive snapshots of the system state, separated by fixed intervals $\Delta t$. We restrict our experiments to static conditioning fields, but the formulation extends directly to time-varying fields, $\mathbf{C}(x,y,t)$, by computing a set of modulation parameters at each input timestep (Sec.~\ref{sec:conclusion}). We consider both $\mathbf{u}$ and $\mathbf{C}$ fields to be discretized on an $H \times W$ spatial grid. Each input snapshot is partitioned into a collection of $N_x \times N_y$ square patches of size $P \times P$, with $N_x = H/P$ and $N_y = W/P$, which are each embedded into the model's latent space of embedding dimension $D$.

    \subsection{Adaptive LayerNorm}
    \label{subsec:adaln}

        Rather than learning static gain and bias parameters as in standard Layer Normalization~\citep{ba2016layernormalization}, \texttt{AdaLN}, as introduced in the Diffusion Transformer~\citep{peebles2023dit}, predicts the scale and shift from an external conditioning embedding $\mathbf{c} \in \mathbb{R}^{D_c}$.

        Formally, let $\mathbf{h} \in \mathbb{R}^{D}$ denote the embedding of a single patch at a single timestep. We project the conditioning vector $\mathbf{c}$ via a linear layer to obtain token-independent scale, shift, and gate vectors, denoted as $\boldsymbol{\gamma}(\mathbf{c}), \boldsymbol{\beta}(\mathbf{c}), \boldsymbol{\alpha}(\mathbf{c}) \in \mathbb{R}^{D}$, respectively:
        \begin{equation}
            \boldsymbol{\gamma}(\mathbf{c}) = \mathbf{W}_\gamma \mathbf{c} + \mathbf{b}_\gamma, \quad \boldsymbol{\beta}(\mathbf{c}) = \mathbf{W}_\beta \mathbf{c} + \mathbf{b}_\beta, \quad \boldsymbol{\alpha}(\mathbf{c}) = \mathbf{W}_\alpha \mathbf{c} + \mathbf{b}_\alpha,
        \end{equation}
        where $\mathbf{W}_\gamma, \mathbf{W}_\beta, \mathbf{W}_\alpha \in \mathbb{R}^{D \times D_c}$ and $\mathbf{b}_\gamma, \mathbf{b}_\beta, \mathbf{b}_\alpha \in \mathbb{R}^{D}$ are learnable projection weights and biases. The scale and shift then give the adaptive normalized representation
        \begin{equation}
            \texttt{AdaLN}(\mathbf{h}, \mathbf{c}) = \left(\mathbf{1} + \boldsymbol{\gamma}(\mathbf{c})\right) \odot \texttt{LayerNorm}(\mathbf{h}) + \boldsymbol{\beta}(\mathbf{c}),
        \end{equation}
        where $\mathbf{1}$ represents a vector of ones, while the gate $\boldsymbol{\alpha}(\mathbf{c})$ scales the sublayer output before it re-enters the residual stream, as made explicit in Sec.~\ref{subsec:padaln}.

    \subsection{Patchwise Adaptive Normalization}
      \label{subsec:padaln}

        \begin{wrapfigure}{r}{0.48\textwidth}
            \centering
            \resizebox{\linewidth}{!}{\input{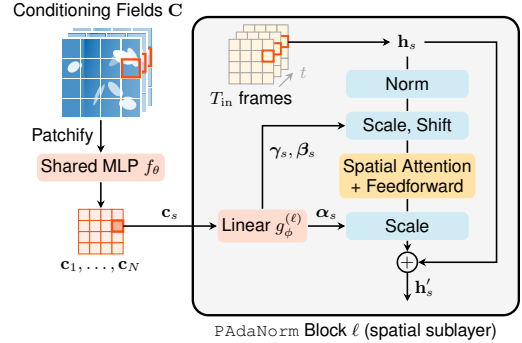}\unskip}%
            \caption{\textbf{Patchwise Adaptive Normalization}. The spatially varying conditioning fields $\mathbf{C}$ are partitioned into patches aligned with the input tokenization. Each patch is independently mapped through a shared MLP $f_\theta$ to produce a local conditioning vector $\mathbf{c}_s$. Each token $\mathbf{h}_s$, in every frame of the history window, is then modulated by the scale, shift, and gate derived from $\mathbf{c}_s$ (conditioning vector of the corresponding patch).}
            \label{fig:padaln}
        \end{wrapfigure}

        The above-described standard implementation of AdaLN applies the same modulation uniformly to all spatial tokens via a single global conditioning vector $\mathbf{c}$. This is appropriate when the conditioning signal is spatially homogeneous (e.g., a scalar timestep or a global parameter), but limiting when the conditioning fields exhibit spatial structure, e.g., heterogeneous material properties (density, refractive index, or heat capacity).

        We introduce \emph{Patchwise Adaptive Normalization} (\texttt{PAdaNorm}), which produces spatially varying modulation parameters by operating at the granularity of spatial patches. Given $C_{\mathrm{cond}}$ spatially varying conditioning fields stacked as $\mathbf{C}(x,y) \in \mathbb{R}^{C_{\mathrm{cond}} \times H \times W}$, we partition them into the same $N_x \times N_y$ grid of non-overlapping patches of size $P \times P$ used for tokenizing the input state. Let $s = (i,j) \in [1, N_x] \times [1, N_y]$ index a patch, $\mathcal{P}_s$ denote its $P{\times}P$ pixel region, and $\mathbf{C}|_{\mathcal{P}_s} \in \mathbb{R}^{C_{\mathrm{cond}} \times P \times P}$ the sub-tensor of the conditioning fields restricted to that region. Each patch is flattened into a vector of dimension $C_{\mathrm{cond}} P^2$ and mapped through a shared encoder $f_\theta$ to produce a local conditioning vector:
        \begin{equation}
          \mathbf{c}_s = f_\theta\!\left(\text{vec}\!\left(\mathbf{C}|_{\mathcal{P}_s}\right)\right) \in \mathbb{R}^{D_c},
        \end{equation}
        where $\text{vec}(\cdot)$ denotes the flattening operator. In practice, $f_\theta$ is an MLP with hidden width $D_h$, output dimension $D_c$, and SiLU activations. $f_\theta$ is shared across all patches and all blocks, and evaluated once per forward pass. This yields a set of patch-local conditioning vectors $\{\mathbf{c}_s\}_{s \in [1, N_x] \times [1, N_y]}$, each depending only on the conditioning values within its corresponding patch.

        Each block $\ell$ then maps these conditioning vectors to per-patch scale, shift, and gate parameters through its own projection $g^{(\ell)}_\phi$, a single linear layer:
        \begin{equation}
            \boldsymbol{\gamma}^{(\ell)}_s,\; \boldsymbol{\beta}^{(\ell)}_s,\; \boldsymbol{\alpha}^{(\ell)}_s = g^{(\ell)}_\phi(\mathbf{c}_s) \in \mathbb{R}^{D} \times \mathbb{R}^{D} \times \mathbb{R}^{D}.
        \end{equation}
        These modulate the single normalization and the single residual gate of an attention or feed-forward block; our main implementation modulates only the spatial attention block (\emph{spatial-only} configuration, Fig.~\ref{fig:padaln}), although the modulation can be placed on any block that is preceded by a normalization and followed by a residual gate (App.~\ref{app:methods}). Writing $\mathbf{h}_s$ for the token embedding at spatial position $s$ entering block $\ell$, and $\tilde{\mathbf{H}}$ for the collection of modulated tokens at a given history timestep, we define:
        \begin{equation}
        \begin{aligned}
                \tilde{\mathbf{h}}_s &= \left(\mathbf{1} + \boldsymbol{\gamma}^{(\ell)}_s\right) \odot \texttt{Norm}(\mathbf{h}_s) + \boldsymbol{\beta}^{(\ell)}_s, \\[2pt]
                \mathbf{h}'_s &= \mathbf{h}_s
                + \boldsymbol{\alpha}^{(\ell)}_s \odot
 \left[
                  \texttt{Attn}(\tilde{\mathbf{H}})_s
                  + \texttt{FFN}(\tilde{\mathbf{h}}_s)
 \right],
        \end{aligned}
        \end{equation}
        where $\texttt{Attn}(\cdot)$ and $\texttt{FFN}(\cdot)$ are the two parallel branches of the spatial attention block (Eq.~\ref{eq:spatial_block}), $\texttt{Norm}$ is a normalization layer (e.g. \texttt{LayerNorm} or \texttt{RMSGroupNorm}), and $\mathbf{1}$ is a vector of ones.

        Following the AdaLN-Zero strategy, each $g^{(\ell)}_\phi$ is initialized so that every block starts close to the identity and early training remains stable. Also, because the conditioning fields we consider are static, they are identical at every timestep of the input window, so a single set of modulation parameters $(\boldsymbol{\gamma}^{(\ell)}_s, \boldsymbol{\beta}^{(\ell)}_s, \boldsymbol{\alpha}^{(\ell)}_s)$ per block is shared by all $T_{\mathrm{in}}$ timesteps.

        Crucially, $f_\theta$ is shared across all spatial patches \emph{and} all blocks, while each block owns a separate projection $g^{(\ell)}_\phi$; Appendix~\ref{app:methods} reports the variants that share $g_\phi$ across blocks as well. The number of conditioning parameters is therefore \textit{independent} of the number of spatial patches and grows only linearly with depth; the conditioning compute, unlike cross-attention, grows linearly with the number of patches and does not depend on the history length $T_{\mathrm{in}}$. A single evaluation of $f_\theta$ and $g^{(\ell)}_\phi$ can therefore be cached and reused through the entire rollout; App.~\ref{app:flops} derives the cost of each conditioning strategy and reports measured FLOP counts.

        In our default implementation, conditioning patches are non-overlapping and aligned one-to-one with state patches, so each token $\mathbf{h}_s$ receives the conditioning vector $\mathbf{c}_s$ of its own patch. \texttt{PAdaNorm} only requires, however, that every token be assigned one conditioning vector: a vector can be shared by several neighboring tokens (coarse patches) or computed from a window that overlaps neighboring patches, both of which we study in Sec.~\ref{subsec:overlapping_patches}. At the other extreme, if a single conditioning vector $\mathbf{c}$ is shared by all tokens, \texttt{PAdaNorm} reduces to the global AdaNorm baseline of App.~\ref{app:adaln_baseline}.

\section{Experiments}
\label{sec:experiments}

    \subsection{Experimental Settings}
    \label{subsec:exper_settings}

        \paragraph{Datasets}
            We evaluate on five public datasets from three sources. From The Well~\citep{thewelldataset2024}, we use three acoustic scattering datasets; from FlowBench~\citep{tali2024flowbench}, the Flow Past Objects dataset; and CO\textsubscript{2}--water flow in porous media from \citet{abdellatif2025co2pore}.
            \textit{i) Acoustic scattering (The Well):} the Maze, Inclusions, and Discontinuous datasets simulate acoustic wave propagation through heterogeneous media at $256{\times}256$ resolution. State variables are pressure and velocity (3 channels), and one-way coupled conditioning fields are density and speed of sound (2 channels), whose spatial structure defines the scattering geometry. In the \textit{Joint} setting, a single model is trained on all three acoustic datasets simultaneously.
            \textit{ii) Flow Past Objects (FlowBench):} 2D Navier--Stokes simulations at $1024{\times}256$ resolution, where state fields are the two velocity components and pressure (3 channels), and conditioning channels are the obstacle binary mask, signed distance function, and Reynolds number.
            \textit{iii) CO\textsubscript{2}--water flow in porous media:} simulates supercritical CO\textsubscript{2} injected into a water-saturated porous rock at the pore scale. State fields are the water saturation, pressure, and the two $x$ and $y$ velocity components; conditioning channels are the binary pore mask separating pore space from rock grains and two normalized spatial coordinate channels. This setting is complementary to the two above in that a) it is a different PDE family, and b) its physics are considerably more complex, as the CO\textsubscript{2} front advances along the path of least resistance through the rock, so information across the entire front must be aggregated to determine where it progresses next.

        \paragraph{Benchmark}
            We compare \texttt{PAdaNorm} against several conditioning strategies. The \textit{In-context} baselines concatenate the conditioning fields to the input state as additional channels before patch embedding, as in Walrus~\citep{mccabe2025walruscrossdomainfoundationmodel}. To control for the parameter overhead introduced by the \texttt{PAdaNorm} conditioning MLPs, we include parameter-matched In-context variants that increase either model width or depth to match the \texttt{PAdaNorm} parameter count. We also compare against a global \textit{AdaNorm} baseline~(see App.~\ref{app:adaln_baseline}), which average-pools the conditioning fields and maps the result, through a single MLP shared across all blocks, to spatially uniform scale, shift, and gate parameters that modulate both attention sublayers.
            We further compare against a (more compute-intensive) \textit{cross-attention} baseline, in which a dedicated encoder tokenizes the conditioning fields on the same patch grid, and all state patches (in space and time) attend to the conditioning patches using spatial 2D RoPE.
            Finally, we include GNOT~\citep{hao2023gnotgeneralneuraloperator}, a neural operator transformer baseline that uses linear attention with heterogeneous cross-attention for conditioning.
            All models are evaluated via 60-step autoregressive rollouts. We report the variance-scaled RMSE (vRMSE)~\citep{thewelldataset2024}, the RMSE normalized by the spatial standard deviation of the ground truth,
            \begin{equation}
                \mathrm{vRMSE}(\mathbf{u}, \hat{\mathbf{u}}) = \left(\frac{\langle |\mathbf{u} - \hat{\mathbf{u}}|^2 \rangle}{\langle |\mathbf{u} - \bar{\mathbf{u}}|^2 \rangle + \epsilon}\right)^{1/2},
                \label{eq:vrmse}
            \end{equation}
            where $\mathbf{u}$ is the ground truth, $\hat{\mathbf{u}}$ the prediction, $\langle \cdot \rangle$ the spatial mean, $\bar{\mathbf{u}} = \langle \mathbf{u} \rangle$, and $\epsilon = 10^{-7}$.
            We evaluate Eq.~\ref{eq:vrmse} and average over channels and validation windows to obtain one value per rollout step, and then aggregate into three horizons: first step $T{=}1$, short $T{\in}[1{:}20]$, and long horizon $T{\in}[21{:}60]$. For each run, we select the checkpoint with the lowest $T{\in}[21{:}60]$ validation vRMSE and report all metrics on the same validation windows, with the compute budget fixed in advance at 50,000 steps.
        \begingroup
\newcommand{\tabmeanstd}[2]{\begin{tabular}[t]{@{}c@{}}$#1$\\[-1pt]{\scriptsize\color{black!60}$\pm #2$}\end{tabular}}
\begin{table}[t!]
\centering
\caption{%
  Long-horizon validation vRMSE ($\downarrow$, $T{\in}[21{:}60]$) for baselines and
  \texttt{PAdaNorm} spatial-only, reported as mean $\pm$ standard deviation over three
  seeds.
  Best mean in \textbf{bold} and second best \underline{underlined} within
  each dataset. $^\ddagger$Parameter-matched In-context controls.
  The \emph{Joint} result reports vRMSE averaged over the three datasets, for a single model trained on all three simultaneously.
  GFLOPs denotes one forward pass at batch size one; cached rollout costs are
  reported in Table~\ref{tab:flops}.
}
\label{tab:main_aggregated}
\small
\setlength{\tabcolsep}{2pt}
\renewcommand{\arraystretch}{1.12}
\oursbox[19pt][14.5pt]{ours-main}
\begin{tabular*}{\textwidth}{@{\extracolsep{\fill}}l*{7}{c}@{}}
  \toprule
  & \shortstack{In-context\\(base)}
  & \shortstack{In-context\\(wide)$^\ddagger$}
  & \shortstack{In-context\\(deep)$^\ddagger$}
  & \shortstack{AdaNorm\\(global)}
  & \shortstack{Cross-\\attention}
  & \shortstack{GNOT\\\phantom{}}
  & \oursL{ours-main}\shortstack{\texttt{PAdaNorm}\\(ours)}\oursR{ours-main} \\
  \midrule
  \multicolumn{8}{@{}l}{\emph{Acoustic scattering (The Well~\citep{thewelldataset2024})}} \\
  \#Params
    & 13.3\,M & 18.0\,M & 18.6\,M & 15.5\,M & 18.5\,M & 18.8\,M & 19.1\,M \\
  GFLOPs
    & 113.5 & 138.9 & 158.6 & 113.3 & 179.7 & 41.2 & 125.2 \\
  \addlinespace[3pt]
  \href{https://polymathic-ai.org/the_well/datasets/acoustic_scattering_maze/}{Maze}
    & \tabmeanstd{0.4917}{0.0226}
    & \tabmeanstd{\underline{0.4101}}{0.0171}
    & \tabmeanstd{0.4522}{0.0470}
    & \tabmeanstd{0.4836}{0.0466}
    & \tabmeanstd{0.5267}{0.0356}
    & \tabmeanstd{1.4555}{0.0845}
    & \tabmeanstd{\mathbf{0.3441}}{0.0288} \\
  \addlinespace[1.5pt]
  \href{https://polymathic-ai.org/the_well/datasets/acoustic_scattering_inclusions/}{Inclusions}
    & \tabmeanstd{0.6015}{0.0297}
    & \tabmeanstd{\underline{0.4868}}{0.0666}
    & \tabmeanstd{0.5573}{0.0275}
    & \tabmeanstd{0.8717}{0.0508}
    & \tabmeanstd{0.5929}{0.0381}
    & \tabmeanstd{1.4505}{0.0715}
    & \tabmeanstd{\mathbf{0.4261}}{0.0156} \\
  \addlinespace[1.5pt]
  \href{https://polymathic-ai.org/the_well/datasets/acoustic_scattering_discontinuous/}{Discontinuous}
    & \tabmeanstd{0.4664}{0.0569}
    & \tabmeanstd{\underline{0.3297}}{0.0168}
    & \tabmeanstd{0.4344}{0.0510}
    & \tabmeanstd{0.6521}{0.0416}
    & \tabmeanstd{0.4915}{0.0097}
    & \tabmeanstd{1.3679}{0.0841}
    & \tabmeanstd{\mathbf{0.2890}}{0.0466} \\
  \addlinespace[1.5pt]
  Joint
    & \tabmeanstd{0.9757}{0.3032}
    & \tabmeanstd{\underline{0.8013}}{0.1964}
    & \tabmeanstd{0.8646}{0.2153}
    & \tabmeanstd{0.8792}{0.0736}
    & \tabmeanstd{0.9449}{0.1770}
    & \tabmeanstd{1.4074}{0.0439}
    & \tabmeanstd{\mathbf{0.7376}}{0.1034} \\
  \midrule
  \multicolumn{8}{@{}l}{\emph{\href{https://github.com/baskargroup/FlowBench}{FlowBench FPO}~\citep{tali2024flowbench}}} \\
  \#Params
    & 30.8\,M & 38.6\,M & 39.6\,M & 32.9\,M & 40.2\,M & 36.1\,M & 38.6\,M \\
  GFLOPs
    & 468.2 & 553.7 & 606.1 & 464.6 & 686.0 & 77.1 & 481.2 \\
  \addlinespace[3pt]
  vRMSE
    & \tabmeanstd{0.3755}{0.0029}
    & \tabmeanstd{0.3240}{0.0265}
    & \tabmeanstd{0.3783}{0.0254}
    & \tabmeanstd{\underline{0.3236}}{0.0098}
    & \tabmeanstd{0.3431}{0.0353}
    & \tabmeanstd{0.3959}{0.0065}
    & \tabmeanstd{\mathbf{0.3177}}{0.0120} \\
  \midrule
  \multicolumn{8}{@{}l}{\emph{\href{https://doi.org/10.5061/dryad.jm63xsjn5}{CO\textsubscript{2}--water flow in porous media}~\citep{abdellatif2025co2pore}}} \\
  \#Params
    & 29.9\,M & 37.3\,M & 38.8\,M & 32.3\,M & 35.8\,M & 35.3\,M & 37.7\,M \\
  GFLOPs
    & 2005.2 & 2175.5 & 2604.7 & 2002.8 & 3203.0 & 301.2 & 2067.0 \\
  \addlinespace[3pt]
  vRMSE
    & \tabmeanstd{0.7875}{0.0039}
    & \tabmeanstd{0.7966}{0.0072}
    & \tabmeanstd{\underline{0.7847}}{0.0030}
    & \tabmeanstd{\mathbf{0.7785}}{0.0054}
    & \tabmeanstd{0.8027}{0.0141}
    & \tabmeanstd{0.8142}{0.0046}
    & \tabmeanstd{0.7893}{0.0073}\oursB{ours-main} \\
  \bottomrule
\end{tabular*}
\vspace{-0.3cm}
\end{table}
\endgroup

        \paragraph{Hyperparameter settings}
            All models use the Walrus-inspired architecture described in Sec.~\ref{subsec:walrus} and are trained for 50{,}000 steps with AdamW and a per-field $L_1$ loss, except on CO\textsubscript{2}, which uses an $L_2$ loss. On the acoustic datasets, models use a patch size of 8, an embedding dimension of 256, and 10 transformer blocks; on FlowBench, a patch size of 16 with an embedding dimension of 384; on CO\textsubscript{2}, a patch size of 8 with an embedding dimension of 384 and 10 blocks, with four input timesteps. The \texttt{PAdaNorm} conditioning encoder is a 4-layer MLP with hidden dimension 512, shared across patches and blocks. Full configuration details, including parameter-matched model sizes, are provided in App.~\ref{app:experimental-settings}.

    \subsection{Results}
    \label{subsec:results}

        \paragraph{Comparison with in-context conditioning}
            At comparable parameter counts, we obtain the clearest accuracy gains on the acoustic datasets by allocating capacity to \texttt{PAdaNorm}'s spatial conditioning. We compare the default spatial-only configuration with wider and deeper In-context models that receive the conditioning fields as additional input channels~(Table~\ref{tab:main_aggregated}). Across three seeds, we reduce mean long-horizon vRMSE by 8--16\% relative to the strongest parameter-matched In-context baseline on each acoustic dataset and in the Joint setting, where a single model is trained on all three. The improvement therefore comes from the dedicated conditioning pathway rather than from the additional parameters it introduces. On FlowBench, we obtain lower mean errors than the In-context baselines at all three reported horizons, although the long-horizon differences are small relative to the seed variation~(Table~\ref{tab:main_flowbench}). On CO\textsubscript{2}, we find no clear accuracy advantage over input concatenation~(Table~\ref{tab:main_co2}). Representative rollouts on each domain
            are shown in App.~\ref{app:qualitative}.

        \paragraph{Comparison with alternative conditioning mechanisms}
            We next compare \texttt{PAdaNorm} with global adaptive normalization and cross-attention~(Table~\ref{tab:main_aggregated}). We obtain lower mean long-horizon errors than global AdaNorm on the acoustic and FlowBench datasets, with the largest differences on acoustic scattering. In contrast, on CO\textsubscript{2}, global AdaNorm attains the lowest mean long-horizon error. This supports the effectiveness of the patchwise configuration, although the two approaches also differ in modulation placement and parameter count. Compared with cross-attention, we obtain lower mean errors on all five datasets and in the Joint setting. Allowing state tokens to attend over the conditioning fields therefore does not provide a consistent accuracy advantage over \texttt{PAdaNorm}, despite requiring more computation. We also obtain lower long-horizon errors than GNOT, which uses a different architecture. We examine computational costs below and compare spatial-only conditioning with other \texttt{PAdaNorm} variants in the ablations.

        \paragraph{Hard constraints on CO\textsubscript{2}}
            Although long-horizon errors on CO\textsubscript{2} are close across models (Table~\ref{tab:main_co2}), the dataset reveals a difference that vRMSE alone does not capture. It allows us to test how well each conditioning mechanism enforces a \emph{hard constraint}: CO\textsubscript{2} and water should flow \emph{around} the impermeable rock grains, never into them. Measuring the mean absolute error on rock pixels alone (Tab.~\ref{tab:rock_mae_co2}) shows that \texttt{PAdaNorm} is best at enforcing this constraint, with the lowest error on all four fields on average across the three seeds. Interestingly, global AdaNorm ranks second on three of the four fields, although its modulation carries no spatial information, suggesting that part of the advantage stems from injecting the conditioning at every block rather than only at the input.

        \paragraph{Computational efficiency}
            Across all datasets, we achieve improved or competitive accuracy with consistently lower rollout FLOPs than parameter-matched In-context baselines and cross-attention models~(Table~\ref{tab:flops}, App.~\ref{app:flops}). Because the conditioning fields we consider are static, \texttt{PAdaNorm}'s conditioning embeddings and modulation parameters can be computed once and reused throughout inference.
            Cross-attention can also reuse the conditioning embeddings and their key and value projections, but must recompute the state-dependent queries, attention products, and output projections. Accounting for reusable computations in both methods, we use 29--36\% fewer FLOPs than cross-attention over a 60-step rollout. The computational benefit of \texttt{PAdaNorm} is not limited to caching: even when recomputing the conditioning parameters, we require 30--35\% fewer forward-pass FLOPs than the evaluated cross-attention configurations. With caching, \texttt{PAdaNorm}'s total rollout cost remains approximately equal to that of the original, smaller In-context model. We therefore improve accuracy on acoustics and FlowBench at little additional rollout cost, while on CO\textsubscript{2} we achieve comparable accuracy with less computation than the wider and deeper controls and cross-attention.

\section{Discussion}
\label{sec:discussion}

    \subsection{What makes \texorpdfstring{\texttt{PAdaNorm}}{PAdaNorm} conditioning effective?}

        \begin{table}[ht!]
\centering
\caption{%
  \textbf{\texttt{PAdaNorm} conditioning mode}: long-horizon validation vRMSE ($\downarrow$, $T{\in}[21{:}60]$) on three acoustic scattering datasets from The Well \citep{thewelldataset2024}.
  We highlight the best metric in \textbf{bold} and \underline{underline} the second best.
  The \emph{Joint} column reports vRMSE averaged over the three datasets, each contributing 200 validation windows, for a single model trained on all three simultaneously.
  All entries are single runs (seed 1234); the three-seed results for spatial-only are in Tab.~\ref{tab:main_aggregated}.
}
\label{tab:ablation_padaln_aggregated}
\small
\begin{tabular}{
    l
    r
    c!{\hskip 4mm}
    c!{\hskip 4mm}
    c!{\hskip 4mm}
    c
}
\toprule
Conditioning
  & \#Params
  & {\href{https://polymathic-ai.org/the_well/datasets/acoustic_scattering_maze/}{Maze}}
  & {\href{https://polymathic-ai.org/the_well/datasets/acoustic_scattering_inclusions/}{Inclusions}}
  & {\href{https://polymathic-ai.org/the_well/datasets/acoustic_scattering_discontinuous/}{Discontinuous}}
  & {Joint}
  \\
\midrule
Spatial-only
  & 19.1\,M
  & 0.3716
  & \textbf{0.4094}
  & \textbf{0.2352}
  & 0.8015
  \\
Temporal-only
  & 19.1\,M
  & \textbf{0.2996}
  & 0.7522
  & 0.3864
  & 0.9520
  \\
Temporal (shared proj.)
  & 19.1\,M
  & 0.3352
  & 0.9115
  & 0.4749
  & 1.0266
  \\
Temporal + Spatial
  & 23.1\,M
  & 0.3203
  & 0.7745
  & 0.3732
  & 0.9325
  \\
Fully shared
  & 15.6\,M
  & 0.3447
  & \underline{0.4819}
  & 0.3553
  & \underline{0.7899}
  \\
Fully shared (spatial)
  & 15.6\,M
  & 0.3522
  & 0.5263
  & \underline{0.2657}
  & 0.8628
  \\
Fully shared (temporal)
  & 15.6\,M
  & \underline{0.3123}
  & 0.5745
  & 0.3864
  & 0.9823
  \\
Shared per-axis
  & 16.9\,M
  & 0.3360
  & 0.5525
  & 0.3199
  & \textbf{0.6726}
  \\
\bottomrule
\end{tabular}
\vspace{-0.3cm}
\end{table}

        We investigate how best to integrate \texttt{PAdaNorm} within the transformer block by varying three design axes (see Tab.~\ref{tab:padaln_variants} in App.~\ref{app:methods} for the full specification of each variant): (i)~which attention axis is modulated---spatial, temporal, or both; (ii)~whether each block owns a dedicated projection or a single MLP is shared across all blocks; and (iii)~when both axes are modulated, whether they receive axis-specific projections or share one.
        Tab.~\ref{tab:ablation_padaln_aggregated} reports the long-horizon vRMSE (single seed) for all eight variants on the acoustic scattering datasets.

        \paragraph{No single variant dominates}
            The best-performing variant depends on the dataset: temporal-only achieves the lowest error on Maze, spatial-only on Inclusions and Discontinuous, and shared per-axis on Joint.
            Spatial-only nonetheless achieves the lowest error on two settings (Inclusions and Discontinuous), more than any other variant, and is therefore the variant retained in our main comparison (Tab.~\ref{tab:main_aggregated}).

        \paragraph{Which attention axis to modulate}
            Modulating a single attention axis (spatial-only or temporal-only, 19.1\,M) generally matches or outperforms the dual-axis temporal+spatial variant (23.1\,M), despite the latter using 21\% more parameters.
            For instance, on Inclusions, spatial-only reaches 0.4094 versus 0.7745 for temporal+spatial.
            Furthermore, broadcasting a shared projection to both axes (temporal, shared proj.) underperforms both dedicated single-axis variants on three of the four settings, suggesting that forcing a single representation to serve both attention types dilutes the conditioning signal.

        \paragraph{Parameter sharing across blocks}
            Shared variants, which use a single conditioning MLP across all blocks (fully shared and shared per-axis), are competitive with per-block variants despite a lower parameter count (15.6--16.9\,M vs.\ 19.1\,M for spatial-only).
            On individual datasets, they often match per-block performance, e.g., fully shared (temporal) achieves 0.3123 on Maze, close to temporal-only's 0.2996.
            Shared variants even surpass the per-block variant on Joint: shared per-axis reaches 0.6726 and fully shared reaches 0.7899, compared to 0.8015 for spatial-only, with 12\% and 18\% fewer parameters.
            This suggests that per-block specialization is not strictly necessary and that a lightweight, shared conditioning path can be an efficient alternative.

    \subsection{Overlapping patches}
    \label{subsec:overlapping_patches}

        By default, the conditioning fields are partitioned into non-overlapping patches aligned one-to-one with the state patch grid, so each conditioning token captures only the geometry local to its corresponding state patch. To allow each conditioning token to incorporate context from neighboring patches, we consider three alternatives displayed in Fig.~\ref{fig:conditioning_patches}: sliding-window MLP, convolutional encoder and coarse patches. We provide a detailed description of each in App.~\ref{app:different_patching}. The sliding-window MLP includes conditioning patches that overlap their neighbors via a sliding window of size $k > P$, so that a shared MLP sees context beyond the corresponding state patch. The convolutional encoder includes a stack of convolutional layers that process the conditioning fields at full spatial resolution, followed by a strided convolution that downsamples to the patch grid, allowing the receptive field to grow with depth. Finally, Coarse conditioning patches involve partitioning the conditioning fields into patches of size $mP \times mP$, each covering $m \times m$ state patches, and the resulting conditioning vector is broadcast identically to all covered tokens.

        \begin{table}[t!]
            \centering
            \caption{%
              Effect of conditioning patch strategy for spatial-only \texttt{PAdaNorm} on validation vRMSE ($\downarrow$) across three acoustic scattering datasets (60-step autoregressive rollout).
              We compare the standard (base) patching against three alternatives:
              sliding-window overlap, convolutional overlap, and coarse~$2{\times}$.
              All metrics are reported at the checkpoint with the best $T{\in}[21{:}60]$ score.
              We highlight the best metric in \textbf{bold} and \underline{underline} the second best.
              All entries are single runs (seed 1234).
            }
            \label{tab:patching_comparison}
            \small
            \setlength{\tabcolsep}{3pt}
            \begin{tabular}{
    @{}l
    r
    cc!{\hskip 3mm}
    cc!{\hskip 3mm}
    cc@{}
}
    \toprule
    \multirow{2}{*}{Patch strategy}
      & \multirow{2}{*}{\#Params}
      & \multicolumn{2}{c!{\hskip 3mm}}{\href{https://polymathic-ai.org/the_well/datasets/acoustic_scattering_maze/}{Maze}}
      & \multicolumn{2}{c!{\hskip 3mm}}{\href{https://polymathic-ai.org/the_well/datasets/acoustic_scattering_inclusions/}{Inclusions}}
      & \multicolumn{2}{c}{\href{https://polymathic-ai.org/the_well/datasets/acoustic_scattering_discontinuous/}{Discontinuous}}
      \\
    \cmidrule(lr{3mm}){3-4}\cmidrule(lr{3mm}){5-6}\cmidrule(lr){7-8}
    & & {$T{\in}[21{:}60]$} & {Overall}
      & {$T{\in}[21{:}60]$} & {Overall}
      & {$T{\in}[21{:}60]$} & {Overall}
      \\
    \midrule
    Base
      & 19.1\,M
      & \underline{0.3716} & \underline{0.2938}
      & 0.4094 & 0.3054
      & \underline{0.2352} & \underline{0.1732}
      \\
    Sliding-window overlap
      & 19.1\,M
      & 0.3770 & 0.2982
      & \textbf{0.3461} & \textbf{0.2629}
      & \textbf{0.2291} & \textbf{0.1688}
      \\
    Convolutional overlap$^*$
      & 20.8\,M
      & \textbf{0.2927} & \textbf{0.2236}
      & \underline{0.4076} & \underline{0.3037}
      & 0.2511 & 0.1852
      \\
    Coarse $2{\times}$
      & 19.3\,M
      & 0.4457 & 0.3374
      & 0.5770 & 0.4311
      & 0.3566 & 0.2604
      \\
    \bottomrule
\end{tabular}
\vspace{-.5cm}

        \end{table}

        We evaluate all three strategies on the acoustic scattering datasets with the spatial-only \texttt{PAdaNorm} conditioning (Tab.~\ref{tab:patching_comparison}).
        The sliding-window overlap improves over the non-overlapping baseline on two of three datasets at no additional parameter cost, with the largest gain on Inclusions ($-15\%$ in long-horizon vRMSE).
        The convolutional encoder achieves the best result on Maze ($-21\%$), is on par with the baseline on Inclusions, and is worse on Discontinuous ($+7\%$), despite a moderate increase in parameters.
        Coarse $2{\times}$ patches degrade performance on all three datasets.
        These results are consistent with the physics of acoustic scattering, where the most important features are material interfaces, at which waves are reflected and diffracted. When such an interface falls on a patch edge, non-overlapping patches split it across two tokens, each missing the full local geometry. Overlapping strategies mitigate this by providing each conditioning token with context from neighboring patches; conversely, coarsening trades spatial resolution for receptive field size, losing the fine-grained geometric detail that drives accurate scattering predictions.

\section{Conclusion}
\label{sec:conclusion}

We investigated how explicitly conditioning on spatial inputs, such as material properties and obstacle geometry, affects transformer-based PDE surrogates. We introduced Patchwise Adaptive Normalization (\texttt{PAdaNorm}) for exogenous fields that influence the evolving state. Through a shared encoder and block-specific projections, we map these fields to spatially aligned scale, shift, and gate parameters that modulate the transformer's intermediate representations.

Across three physical domains, this conditioning approach improves or matches prediction accuracy while reducing rollout computation relative to similarly sized baselines. On acoustic scattering, we obtain reductions of up to 16\% in mean long-horizon variance-scaled RMSE compared with parameter-matched baselines that concatenate conditioning fields with the input state. Across all benchmarks, we obtain improved or competitive accuracy with fewer rollout FLOPs than the wider and deeper models or the cross-attention baselines. \texttt{PAdaNorm} also better maintains hard physical constraints over long rollouts, such as preventing flow into impenetrable solid regions in porous media. Further, since the conditioning fields considered are static, modulation parameters can be cached once and reused throughout inference, amortizing the associated computational cost. Ablations further show that preserving distinct conditioning vectors at the patch resolution is beneficial, while additional spatial context and modulation complexity have dataset-dependent effects. Taken together, the spatial conditioning provided by \texttt{PAdaNorm} is an effective use of model capacity, improving prediction accuracy with fewer forward-pass FLOPs than parameter-matched baselines (App.~\ref{app:flops}).
\vspace{-.25cm}

\paragraph{Limitations and future work}
    Our evaluation is restricted to two-dimensional PDEs and a single backbone architecture (Walrus); validating \texttt{PAdaNorm} on three-dimensional domains and other transformer-based surrogates remains open.
    All conditioning fields considered in this work are static; extending \texttt{PAdaNorm} to time-varying exogenous fields, where the conditioning must be recomputed at each rollout step, is a natural next step.
    The optimal \texttt{PAdaNorm} variant (spatial-only vs.\ temporal-only) is dataset-dependent, and we currently lack a principled criterion for selecting the modulation axis a priori.
    Finally, while we empirically observe that \texttt{PAdaNorm} lowers long-horizon errors on most of the datasets considered and better maintains hard physical constraints over long rollouts, a theoretical understanding of why spatially localized modulation has these effects would further strengthen the approach.

\bibliography{iclr2027_conference}
\bibliographystyle{iclr2027_conference}

\newpage
\appendix
\section{Model details}
    \subsection{Walrus Transformer}
        We detail our Walrus-inspired architecture here, which retains the spatial and temporal attention blocks of the original Walrus backbone~\citep{mccabe2025walruscrossdomainfoundationmodel}.

    \label{app:subsec:walrus_transformer}
        \paragraph{Temporal attention}
            In the temporal attention block, each spatial patch attends only to patches at the same spatial location across the timesteps within the input window. Formally, given an input spatiotemporal feature tensor $\mathbf{X} \in \mathbb{R}^{B \times T_{\mathrm{in}} \times N_x \times N_y \times D}$, where $B$ is the batch size, $T_{\mathrm{in}}$ is the number of context timesteps, and $D$ is the embedding dimension, the temporal block performs multi-head self-attention (MHSA) independently across the time axis for each spatial token. For a patch position $s \in [1, N_x]\times [1,N_y]$, let $\mathbf{X}_s \in \mathbb{R}^{B \times T_{\mathrm{in}} \times D}$ denote the sub-tensor of $\mathbf{X}$ at that position and $\tilde{\mathbf{X}}_s = \texttt{RMSGroupNorm}(\mathbf{X}_s)$, where \texttt{RMSGroupNorm} partitions the $D$ channels into $N_H$ groups ($N_H$ being the number of attention heads), applies RMS normalization within each group and rescales by a learned per-channel gain. The queries, keys, and values are mapped via projection weights $\mathbf{W}_q, \mathbf{W}_k, \mathbf{W}_v \in \mathbb{R}^{D \times D}$, so that the updated representation $\mathbf{X}'_s$ reads:
            \begin{equation}
                \mathbf{X}'_s = \mathbf{X}_s + \texttt{Softmax}\left(\frac{(\tilde{\mathbf{X}}_s\mathbf{W}_q)(\tilde{\mathbf{X}}_s\mathbf{W}_k)^\top}{\sqrt{d_k}}\right)(\tilde{\mathbf{X}}_s\mathbf{W}_v)\mathbf{W}_o,
            \end{equation}
            where $\mathbf{W}_o \in \mathbb{R}^{D \times D}$ is the output projection matrix and $d_k$ is the per-head scaling dimension; per-head query--key normalization and the 1D temporal RoPE are applied to the queries and keys, and are omitted for simplicity of notation. As noted in Sec.~\ref{sec:method}, this attention is unmasked: all $T_{\mathrm{in}}$ frames of the input window are observed, so the block acts as a bidirectional encoder over the context, while the prediction itself remains causal by construction since the target frame lies outside the window. The temporal sublayer carries no feed-forward block, as in the original Walrus implementation. The full intermediate tensor $\mathbf{X}' \in \mathbb{R}^{B \times T_{\mathrm{in}} \times N_x \times N_y \times D}$ is reconstructed by stacking $\mathbf{X}'_s$ over spatial positions.

        \paragraph{Spatial attention}
            Following the temporal update, the intermediate tensor $\mathbf{X}'$ is forwarded to the spatial attention block, which allows each token to attend to all other positions at the same time instant. Unlike the temporal sublayer, this block carries out the layer's feed-forward computation. It does so in \emph{parallel} with the attention rather than in sequence: following the spatial attention design of Walrus~\citep{mccabe2025walruscrossdomainfoundationmodel}, a single normalization feeds one fused projection that produces the attention queries, keys, and values together with the gate and up projections of a SwiGLU feed-forward branch, and the two branch outputs are summed into a single residual.

            For a fixed timestep $t \in [1, T_{\mathrm{in}}]$, let $\mathbf{X}'_t = \mathbf{X}'_{:, t, :, :} \in \mathbb{R}^{B \times N_x \times N_y \times D}$ and $\tilde{\mathbf{X}}'_t = \texttt{RMSGroupNorm}(\mathbf{X}'_t)$. The fused projection $\mathbf{W}_{\mathrm{fused}} \in \mathbb{R}^{D \times (3D + 2D_{\mathrm{ff}})}$ yields
            \begin{equation}
                \left[\mathbf{Q}, \mathbf{K}, \mathbf{V}, \mathbf{G}, \mathbf{U}\right] = \tilde{\mathbf{X}}'_t\mathbf{W}_{\mathrm{fused}},
            \end{equation}
            with $\mathbf{Q}, \mathbf{K}, \mathbf{V} \in \mathbb{R}^{B \times N_xN_y \times D}$ and $\mathbf{G}, \mathbf{U} \in \mathbb{R}^{B \times N_xN_y \times D_{\mathrm{ff}}}$, where $D_{\mathrm{ff}}$ is the feed-forward width. Omitting the 2D RoPE applied to $\mathbf{Q}$ and $\mathbf{K}$, the output representation $\mathbf{X}''_t$ reads:
            \begin{equation}
                \mathbf{X}''_t = \mathbf{X}'_t
                  + \underbrace{\texttt{Softmax}\left(\frac{\mathbf{Q}\mathbf{K}^\top}{\sqrt{d_k}}\right)\mathbf{V}\mathbf{W}'_o}_{\text{attention branch}}
                  + \underbrace{\left(\texttt{SiLU}(\mathbf{G}) \odot \mathbf{U}\right)\mathbf{W}_{\mathrm{out}}}_{\text{feed-forward branch}},
                \label{eq:spatial_block}
            \end{equation}
            where $\mathbf{W}'_o \in \mathbb{R}^{D \times D}$ and $\mathbf{W}_{\mathrm{out}} \in \mathbb{R}^{D_{\mathrm{ff}} \times D}$, and $\odot$ denotes the element-wise product.

\section{Experiments}
\label{app:experiments}

    \subsection{Results}
    \label{app:results}
        Tab.~\ref{tab:conditioning_acoustic} reports the full acoustic results at all three horizons, and Tabs.~\ref{tab:main_flowbench} and~\ref{tab:main_co2} those for FlowBench and CO\textsubscript{2}.
        \begingroup
\newcommand{\tabmeanstd}[2]{\begin{tabular}[t]{@{}c@{}}$#1$\\[-1pt]{\scriptsize\color{black!60}$\pm #2$}\end{tabular}}
\newcommand{\tabsingleseed}[1]{\begin{tabular}[t]{@{}l@{}}#1\\[-1pt]{\scriptsize\color{black!60}(single seed)}\end{tabular}}
\begin{table}[p]
\centering
\caption{%
  Best validation vRMSE ($\downarrow$) across conditioning modes on three acoustic scattering datasets from The Well \citep{thewelldataset2024} (60-step autoregressive rollout).
  All models share similar hyperparameters where relevant.
  Rows that state a spread are the mean $\pm$ sample standard deviation over three seeds
  (1234, 21, 67), each interval read at that seed's best $T{\in}[21{:}60]$ checkpoint;
  rows without one are marked as single-seed and are not directly comparable to them.
  We highlight the best metric in \textbf{bold} and \underline{underline} the second best.
  GFLOPs is the batch-1 forward pass, as in Tab.~\ref{tab:flops}.
  $^\dagger$Parameter-matched In-context controls use the same total parameter count
  as the \texttt{PAdaNorm} temporal+spatial variants (${\sim}$23\,M).
  $^\ddagger$Parameter-matched In-context controls targeting the per-block \texttt{PAdaNorm} parameter count (${\sim}$19\,M).
  The \emph{Joint} column reports vRMSE averaged over the three datasets, each contributing 200 validation windows, for a single model trained on all three simultaneously.
}
\label{tab:conditioning_acoustic}
\small
\setlength{\tabcolsep}{1.5pt}
\resizebox{\linewidth}{!}{%
\begin{tabular}{
    l
    r
    r
    ccc!{\hskip 2mm}
    ccc!{\hskip 2mm}
    ccc!{\hskip 2mm}
    ccc
}
\toprule
\multirow{2}{*}{Conditioning}
  & \multirow{2}{*}{\#Params}
  & \multirow{2}{*}{GFLOPs}
  & \multicolumn{3}{c!{\hskip 2mm}}{\href{https://polymathic-ai.org/the_well/datasets/acoustic_scattering_maze/}{Maze}}
  & \multicolumn{3}{c!{\hskip 2mm}}{\href{https://polymathic-ai.org/the_well/datasets/acoustic_scattering_inclusions/}{Inclusions}}
  & \multicolumn{3}{c!{\hskip 2mm}}{\href{https://polymathic-ai.org/the_well/datasets/acoustic_scattering_discontinuous/}{Discontinuous}}
  & \multicolumn{3}{c}{Joint}
  \\
\cmidrule(lr{2mm}){4-6}\cmidrule(lr{2mm}){7-9}\cmidrule(lr{2mm}){10-12}\cmidrule(lr){13-15}
& & & {$T{=}1$} & {$T{\in}[1{:}20]$} & {$T{\in}[21{:}60]$}
  & {$T{=}1$} & {$T{\in}[1{:}20]$} & {$T{\in}[21{:}60]$}
  & {$T{=}1$} & {$T{\in}[1{:}20]$} & {$T{\in}[21{:}60]$}
  & {$T{=}1$} & {$T{\in}[1{:}20]$} & {$T{\in}[21{:}60]$}
  \\
\midrule
\multicolumn{15}{l}{\emph{Baselines}} \\
In-context
  & 13.3\,M
  & 113.5
  & \tabmeanstd{0.0267}{0.0021} & \tabmeanstd{0.1165}{0.0129} & \tabmeanstd{0.4917}{0.0226}
  & \tabmeanstd{0.0179}{0.0009} & \tabmeanstd{0.1264}{0.0061} & \tabmeanstd{0.6015}{0.0297}
  & \tabmeanstd{0.0084}{0.0001} & \tabmeanstd{0.0807}{0.0058} & \tabmeanstd{0.4664}{0.0569}
  & \tabmeanstd{0.0248}{0.0031} & \tabmeanstd{0.1629}{0.0275} & \tabmeanstd{0.9757}{0.3032}
  \\
\tabsingleseed{In-context$^\dagger$ (wide)}
  & 23.9\,M
  & 184.8
  & 0.0250 & 0.1089 & 0.4566
  & \textbf{0.0140} & 0.1119 & 0.5828
  & 0.0086 & 0.0865 & 0.4255
  & \textbf{0.0179} & 0.1251 & \underline{0.6973}
  \\
\tabsingleseed{In-context$^\dagger$ (deep)}
  & 23.9\,M
  & 203.8
  & \underline{0.0245} & 0.1022 & 0.4472
  & 0.0158 & 0.1172 & 0.6818
  & 0.0076 & 0.0824 & 0.4855
  & \underline{0.0187} & 0.1263 & 0.8152
  \\
In-context$^\ddagger$ (wide)
  & 18.0\,M
  & 138.9
  & \tabmeanstd{\textbf{0.0232}}{0.0004} & \tabmeanstd{0.0987}{0.0050} & \tabmeanstd{0.4101}{0.0171}
  & \tabmeanstd{\underline{0.0141}}{0.0008} & \tabmeanstd{\textbf{0.0976}}{0.0135} & \tabmeanstd{0.4868}{0.0666}
  & \tabmeanstd{\textbf{0.0069}}{0.0005} & \tabmeanstd{0.0583}{0.0019} & \tabmeanstd{0.3297}{0.0168}
  & \tabmeanstd{0.0194}{0.0014} & \tabmeanstd{0.1399}{0.0074} & \tabmeanstd{0.8013}{0.1964}
  \\
In-context$^\ddagger$ (deep)
  & 18.6\,M
  & 158.6
  & \tabmeanstd{0.0292}{0.0012} & \tabmeanstd{0.1178}{0.0102} & \tabmeanstd{0.4522}{0.0470}
  & \tabmeanstd{0.0166}{0.0005} & \tabmeanstd{0.1111}{0.0106} & \tabmeanstd{0.5573}{0.0275}
  & \tabmeanstd{0.0081}{0.0010} & \tabmeanstd{0.0781}{0.0106} & \tabmeanstd{0.4344}{0.0510}
  & \tabmeanstd{0.0215}{0.0021} & \tabmeanstd{0.1571}{0.0446} & \tabmeanstd{0.8646}{0.2153}
  \\
AdaNorm (global)
  & 15.5\,M
  & 113.3
  & \tabmeanstd{0.0328}{0.0030} & \tabmeanstd{0.1856}{0.0120} & \tabmeanstd{0.4836}{0.0466}
  & \tabmeanstd{0.0185}{0.0005} & \tabmeanstd{0.2043}{0.0012} & \tabmeanstd{0.8717}{0.0508}
  & \tabmeanstd{0.0080}{0.0003} & \tabmeanstd{0.1127}{0.0116} & \tabmeanstd{0.6521}{0.0416}
  & \tabmeanstd{0.0241}{0.0033} & \tabmeanstd{0.2031}{0.0100} & \tabmeanstd{0.8792}{0.0736}
  \\
Cross-attention
  & 18.5\,M
  & 179.7
  & \tabmeanstd{0.0249}{0.0009} & \tabmeanstd{0.1129}{0.0103} & \tabmeanstd{0.5267}{0.0356}
  & \tabmeanstd{0.0169}{0.0010} & \tabmeanstd{0.1207}{0.0050} & \tabmeanstd{0.5929}{0.0381}
  & \tabmeanstd{0.0078}{0.0006} & \tabmeanstd{0.0799}{0.0056} & \tabmeanstd{0.4915}{0.0097}
  & \tabmeanstd{0.0220}{0.0009} & \tabmeanstd{0.1541}{0.0146} & \tabmeanstd{0.9449}{0.1770}
  \\
GNOT
  & 18.8\,M
  & 41.2
  & \tabmeanstd{0.2690}{0.0776} & \tabmeanstd{1.0853}{0.0396} & \tabmeanstd{1.4555}{0.0845}
  & \tabmeanstd{0.0943}{0.0289} & \tabmeanstd{0.9829}{0.1135} & \tabmeanstd{1.4505}{0.0715}
  & \tabmeanstd{0.1107}{0.0432} & \tabmeanstd{0.9758}{0.0907} & \tabmeanstd{1.3679}{0.0841}
  & \tabmeanstd{0.2207}{0.0871} & \tabmeanstd{1.0577}{0.1102} & \tabmeanstd{1.4074}{0.0439}
  \\
\midrule
\multicolumn{15}{l}{\oursbox[9.2pt][13.8pt][4.7pt]{ours-acoustic}\oursL{ours-acoustic}\emph{\texttt{PAdaNorm} conditioning} (ours)} \\
Spatial-only
  & 19.1\,M
  & 125.2
  & \tabmeanstd{0.0300}{0.0063} & \tabmeanstd{0.1118}{0.0230} & \tabmeanstd{0.3441}{0.0288}
  & \tabmeanstd{0.0176}{0.0008} & \tabmeanstd{0.1064}{0.0116} & \tabmeanstd{\textbf{0.4261}}{0.0156}
  & \tabmeanstd{0.0077}{0.0003} & \tabmeanstd{\underline{0.0579}}{0.0078} & \tabmeanstd{\underline{0.2890}}{0.0466}
  & \tabmeanstd{0.0242}{0.0028} & \tabmeanstd{\textbf{0.1223}}{0.0055} & \tabmeanstd{0.7376}{0.1034}\oursR{ours-acoustic}
  \\
\tabsingleseed{Temporal-only}
  & 19.1\,M
  & 125.2
  & 0.0270 & \underline{0.0942} & \textbf{0.2996}
  & 0.0172 & 0.1238 & 0.7522
  & 0.0076 & 0.0700 & 0.3864
  & 0.0218 & 0.1249 & 0.9520
  \\
\tabsingleseed{Temporal (shared proj.)}
  & 19.1\,M
  & 125.2
  & 0.0253 & 0.1000 & 0.3352
  & 0.0247 & 0.2026 & 0.9115
  & 0.0074 & 0.0740 & 0.4749
  & 0.0573 & 0.2881 & 1.0266
  \\
\tabsingleseed{Temporal + Spatial}
  & 23.1\,M
  & 133.3
  & 0.0251 & 0.1003 & 0.3203
  & 0.0179 & 0.1215 & 0.7745
  & 0.0074 & 0.0758 & 0.3732
  & 0.0530 & 0.2317 & 0.9325
  \\
\tabsingleseed{Fully shared}
  & 15.6\,M
  & 118.0
  & 0.0260 & \textbf{0.0898} & 0.3447
  & 0.0169 & 0.1030 & \underline{0.4819}
  & 0.0080 & 0.0663 & 0.3553
  & 0.0254 & 0.1558 & 0.7899
  \\
\tabsingleseed{Fully shared (spatial)}
  & 15.6\,M
  & 118.0
  & 0.0698 & 0.1555 & 0.3522
  & 0.0169 & \underline{0.0994} & 0.5263
  & 0.0083 & \textbf{0.0570} & \textbf{0.2657}
  & 0.0218 & 0.1253 & 0.8628
  \\
\tabsingleseed{Fully shared (temporal)}
  & 15.6\,M
  & 118.0
  & 0.0488 & 0.1226 & \underline{0.3123}
  & 0.0174 & 0.1118 & 0.5745
  & \underline{0.0072} & 0.0756 & 0.3864
  & 0.0342 & 0.1842 & 0.9823
  \\
\tabsingleseed{Shared per-axis}
  & 16.9\,M
  & 120.7
  & 0.0271 & 0.1000 & 0.3360
  & 0.0166 & 0.1101 & 0.5525
  & 0.0085 & 0.0692 & 0.3199
  & 0.0224 & \underline{0.1239} & \textbf{0.6726}\oursB{ours-acoustic}
  \\
\bottomrule
\end{tabular}%
}
\end{table}
\endgroup

        \begingroup
\newcommand{\tabmeanstd}[2]{\begin{tabular}[t]{@{}c@{}}$#1$\\[-1pt]{\scriptsize\color{black!60}$\pm #2$}\end{tabular}}
\begin{table}[ht!]
\centering
\caption{%
  Best validation vRMSE ($\downarrow$) for baselines and \texttt{PAdaNorm} spatial-only on
  \href{https://github.com/baskargroup/FlowBench}{FlowBench FPO}~\citep{tali2024flowbench}
  (2D Navier--Stokes, $1024{\times}256$, 60-step autoregressive rollout).
  $^\ddagger$Parameter-matched In-context controls targeting the \texttt{PAdaNorm} parameter
  count (${\sim}38.6$\,M).
  Each entry is the mean $\pm$ sample standard deviation over three seeds (1234, 21, 67).
  Per seed the three intervals are read at that seed's best $T{\in}[21{:}60]$ checkpoint, so each row describes one model per seed.
  GFLOPs is the batch-1 forward pass, reproduced from Tab.~\ref{tab:flops}.
  We highlight the best mean in \textbf{bold} and \underline{underline} the second best.
}
\label{tab:main_flowbench}
\small
\setlength{\tabcolsep}{4pt}
\oursbox[8.5pt][13.5pt]{ours-flowbench}
\begin{tabular}{
    l
    r
    r
    ccc
}
\toprule
Conditioning
  & \#Params
  & GFLOPs
  & {$T{=}1$} & {$T{\in}[1{:}20]$} & {$T{\in}[21{:}60]$}
  \\
\midrule
\multicolumn{6}{l}{\emph{Baselines}} \\
In-context
  & 30.8\,M
  & 468.2
  & \tabmeanstd{0.0014}{0.0000} & \tabmeanstd{0.0455}{0.0022} & \tabmeanstd{0.3755}{0.0029}
  \\
In-context$^\ddagger$ (wide)
  & 38.6\,M
  & 553.7
  & \tabmeanstd{0.0013}{0.0000} & \tabmeanstd{0.0417}{0.0025} & \tabmeanstd{0.3240}{0.0265}
  \\
In-context$^\ddagger$ (deep)
  & 39.6\,M
  & 606.1
  & \tabmeanstd{0.0019}{0.0009} & \tabmeanstd{0.0527}{0.0134} & \tabmeanstd{0.3783}{0.0254}
  \\
AdaNorm (global)
  & 32.9\,M
  & 464.6
  & \tabmeanstd{0.0014}{0.0001} & \tabmeanstd{\underline{0.0402}}{0.0017} & \tabmeanstd{\underline{0.3236}}{0.0098}
  \\
Cross-attention
  & 40.2\,M
  & 686.0
  & \tabmeanstd{\underline{0.0013}}{0.0001} & \tabmeanstd{0.0435}{0.0039} & \tabmeanstd{0.3431}{0.0353}
  \\
GNOT
  & 36.1\,M
  & 77.1
  & \tabmeanstd{0.0026}{0.0001} & \tabmeanstd{0.0651}{0.0017} & \tabmeanstd{0.3959}{0.0065}
  \\
\midrule
\multicolumn{6}{l}{\emph{\texttt{PAdaNorm} conditioning} (ours)} \\
\oursL{ours-flowbench}Spatial-only
  & 38.6\,M
  & 481.2
  & \tabmeanstd{\mathbf{0.0012}}{0.0001} & \tabmeanstd{\mathbf{0.0392}}{0.0036} & \tabmeanstd{\mathbf{0.3177}}{0.0120}\oursR{ours-flowbench}\oursB{ours-flowbench}
  \\
\bottomrule
\end{tabular}
\end{table}
\endgroup

        \begingroup
\newcommand{\tabmeanstd}[2]{\begin{tabular}[t]{@{}c@{}}$#1$\\[-1pt]{\scriptsize\color{black!60}$\pm #2$}\end{tabular}}
\begin{table}[ht!]
\centering
\caption{%
  Best validation vRMSE ($\downarrow$) for baselines and \texttt{PAdaNorm} spatial-only on
  \href{https://doi.org/10.5061/dryad.jm63xsjn5}{CO\textsubscript{2}--water flow in porous media}~\citep{abdellatif2025co2pore}
  ($512{\times}512$, $T_{\mathrm{in}}{=}4$, 60-step autoregressive rollout).
  $^\ddagger$Parameter-matched In-context controls targeting the \texttt{PAdaNorm}
  parameter count (${\sim}37.7$\,M).
  Each entry is the mean $\pm$ sample standard deviation over three seeds (1234, 21, 67).
  Per seed the three intervals are read at that seed's best $T{\in}[21{:}60]$ checkpoint, so each row describes one model per seed.
  GFLOPs is the batch-1 forward pass, reproduced from Tab.~\ref{tab:flops}.
  We highlight the best mean in \textbf{bold} and \underline{underline} the second best.
}
\label{tab:main_co2}
\small
\setlength{\tabcolsep}{4pt}
\oursbox[8.5pt][13.5pt]{ours-co2}
\begin{tabular}{
    l
    r
    r
    ccc
}
\toprule
Conditioning
  & \#Params
  & GFLOPs
  & {$T{=}1$} & {$T{\in}[1{:}20]$} & {$T{\in}[21{:}60]$}
  \\
\midrule
\multicolumn{6}{l}{\emph{Baselines}} \\
In-context
  & 29.9\,M
  & 2005.2
  & \tabmeanstd{0.2652}{0.0029} & \tabmeanstd{0.4703}{0.0064} & \tabmeanstd{0.7875}{0.0039}
  \\
In-context$^\ddagger$ (wide)
  & 37.3\,M
  & 2175.5
  & \tabmeanstd{0.2607}{0.0036} & \tabmeanstd{0.4760}{0.0053} & \tabmeanstd{0.7966}{0.0072}
  \\
In-context$^\ddagger$ (deep)
  & 38.8\,M
  & 2604.7
  & \tabmeanstd{\mathbf{0.2571}}{0.0043} & \tabmeanstd{\underline{0.4682}}{0.0035} & \tabmeanstd{\underline{0.7847}}{0.0030}
  \\
AdaNorm (global)
  & 32.3\,M
  & 2002.8
  & \tabmeanstd{0.2605}{0.0023} & \tabmeanstd{0.4697}{0.0054} & \tabmeanstd{\mathbf{0.7785}}{0.0054}
  \\
Cross-attention
  & 35.8\,M
  & 3203.0
  & \tabmeanstd{0.2644}{0.0013} & \tabmeanstd{0.4781}{0.0135} & \tabmeanstd{0.8027}{0.0141}
  \\
GNOT
  & 35.3\,M
  & 301.2
  & \tabmeanstd{0.3034}{0.0009} & \tabmeanstd{0.5214}{0.0029} & \tabmeanstd{0.8142}{0.0046}
  \\
\midrule
\multicolumn{6}{l}{\emph{\texttt{PAdaNorm} conditioning} (ours)} \\
\oursL{ours-co2}Spatial-only
  & 37.7\,M
  & 2067.0
  & \tabmeanstd{\underline{0.2591}}{0.0027} & \tabmeanstd{\mathbf{0.4666}}{0.0145} & \tabmeanstd{0.7893}{0.0073}\oursR{ours-co2}\oursB{ours-co2}
  \\
\bottomrule
\end{tabular}
\end{table}
\endgroup

        \begingroup
\newcommand{\tabmeanstd}[2]{\begin{tabular}[t]{@{}c@{}}$#1$\\[-1pt]{\scriptsize\color{black!60}$\pm #2$}\end{tabular}}
\begin{table}[ht!]
\centering
\caption{%
  Mean absolute error on \emph{rock} pixels ($\downarrow$, z-score normalized units) on CO\textsubscript{2}--water flow in porous media, averaged over $T{\in}[21{:}60]$.
  All fields must be zero within the rock pixels.
  Each entry is the mean $\pm$ sample standard deviation over three seeds (1234, 21, 67), each read at its best $T{\in}[21{:}60]$ checkpoint, as in Tab.~\ref{tab:main_co2}.
  $^\ddagger$Parameter-matched In-context controls.
  We highlight the best mean in \textbf{bold} and \underline{underline} the second best.
}
\label{tab:rock_mae_co2}
\small
\setlength{\tabcolsep}{6pt}
\oursbox[8.5pt][13.5pt]{ours-rock}
\begin{tabular}{l cccc}
\toprule
Conditioning & Saturation & Pressure & $U_x$ & $U_y$ \\
\midrule
\multicolumn{5}{l}{\emph{Baselines}} \\
In-context & \tabmeanstd{0.083}{0.002} & \tabmeanstd{0.157}{0.023} & \tabmeanstd{0.204}{0.015} & \tabmeanstd{0.224}{0.017} \\
In-context$^\ddagger$ (wide) & \tabmeanstd{0.089}{0.009} & \tabmeanstd{0.224}{0.073} & \tabmeanstd{0.260}{0.026} & \tabmeanstd{0.253}{0.024} \\
In-context$^\ddagger$ (deep) & \tabmeanstd{0.075}{0.006} & \tabmeanstd{\underline{0.155}}{0.007} & \tabmeanstd{0.240}{0.043} & \tabmeanstd{0.239}{0.027} \\
AdaNorm (global) & \tabmeanstd{\underline{0.073}}{0.013} & \tabmeanstd{0.174}{0.010} & \tabmeanstd{\underline{0.182}}{0.015} & \tabmeanstd{\underline{0.192}}{0.013} \\
Cross-attention & \tabmeanstd{0.091}{0.006} & \tabmeanstd{0.193}{0.018} & \tabmeanstd{0.216}{0.018} & \tabmeanstd{0.235}{0.014} \\
GNOT & \tabmeanstd{0.101}{0.023} & \tabmeanstd{0.187}{0.049} & \tabmeanstd{0.227}{0.026} & \tabmeanstd{0.229}{0.011} \\
\midrule
\multicolumn{5}{l}{\emph{\texttt{PAdaNorm} conditioning} (ours)} \\
\oursL{ours-rock}Spatial-only & \tabmeanstd{\mathbf{0.062}}{0.002} & \tabmeanstd{\mathbf{0.114}}{0.002} & \tabmeanstd{\mathbf{0.142}}{0.008} & \tabmeanstd{\mathbf{0.145}}{0.004}\oursR{ours-rock}\oursB{ours-rock} \\
\bottomrule
\end{tabular}
\end{table}
\endgroup

    \subsection{Qualitative rollouts}
    \label{app:qualitative}
\begin{figure}[ht!]
            \centering
            \includegraphics[width=\textwidth]{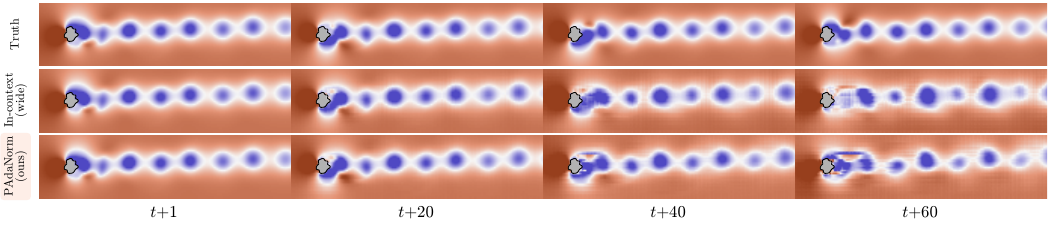}
            \caption{%
              Autoregressive rollout on FlowBench FPO (pressure channel, Reynolds number $\mathrm{Re}{=}364$).
              Rows: ground truth, the parameter-matched wide In-context baseline
              ($D{=}512$, 7 blocks), and \texttt{PAdaNorm} spatial-only ($D{=}384$, 10 blocks); each rollout is generated by the seed-1234 checkpoint from Tab.~\ref{tab:main_aggregated}. The obstacle is overlaid in gray.
            }
            \label{fig:rollout_flowbench}
        \end{figure}

\begin{figure}[ht!]
            \centering
            \includegraphics[width=\textwidth]{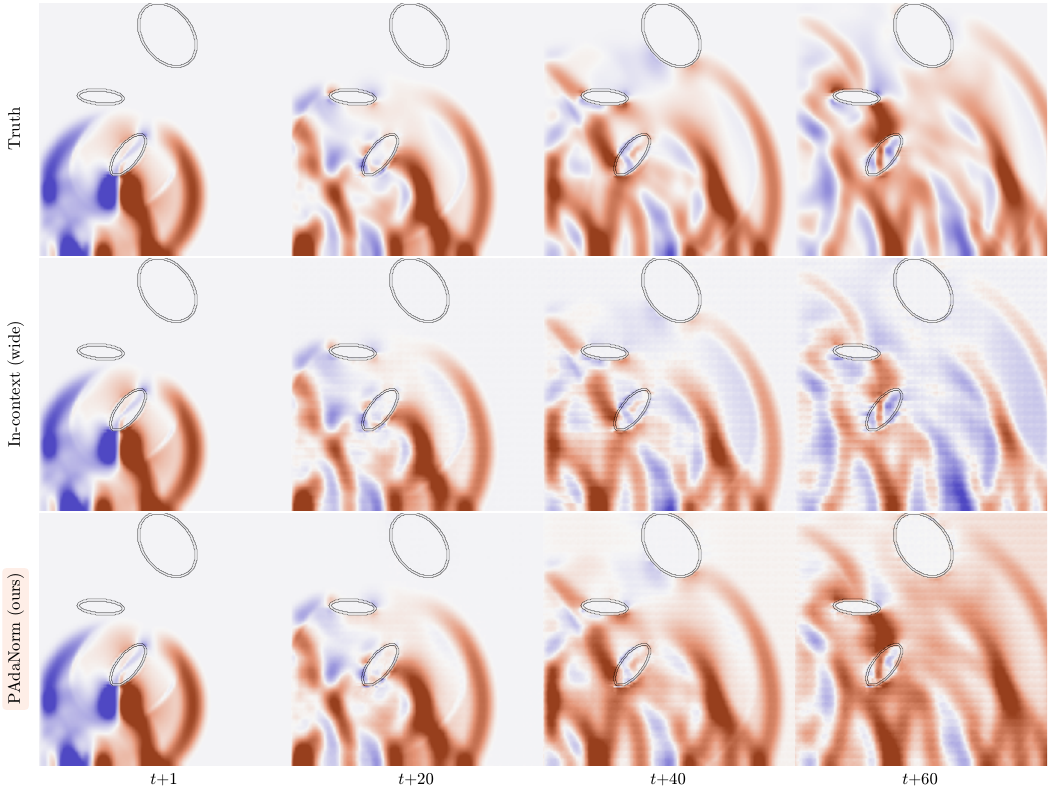}
            \caption{%
              Typical autoregressive rollout on acoustic scattering with inclusions (pressure).
              Rows: ground truth, the parameter-matched wide In-context baseline
              ($D{=}384$, 6 blocks), the strongest In-context baseline on this dataset in Tab.~\ref{tab:main_aggregated}, and \texttt{PAdaNorm} spatial-only ($D{=}256$, 10 blocks); each rollout is generated by a seed-21 checkpoint from the runs whose metrics enter that table.
              The inclusions are outlined in gray. The window is the one whose \texttt{PAdaNorm}-vs-baseline error gap is closest to the median over the candidates considered, so it is neither a best- nor a worst-case draw.
            }
            \label{fig:rollout_inclusions}
        \end{figure}

\begin{figure}[ht!]
            \centering
            \includegraphics[width=\textwidth]{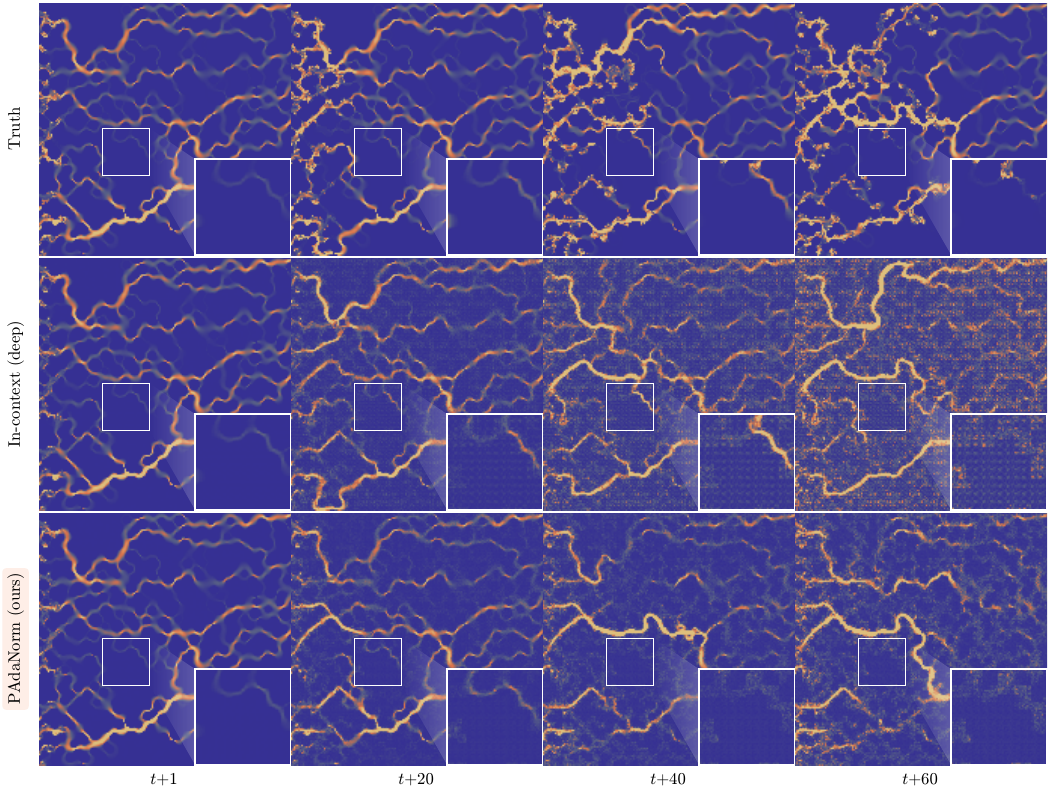}
            \caption{%
              Typical autoregressive rollout on
              \href{https://doi.org/10.5061/dryad.jm63xsjn5}{CO\textsubscript{2}--water flow in porous media}~\citep{abdellatif2025co2pore}, showing the velocity magnitude
              $\lVert\mathbf{u}\rVert$. The inset in the bottom-right corner of each panel magnifies the region outlined in white. The rows show the ground truth, In-context (deep) --- the depth-matched parameter control --- and \texttt{PAdaNorm} spatial-only.
             Throughout the rollout, \texttt{PAdaNorm} enforces the hard constraint of not having gas or water permeating the rock cells better than the parameter-matched In-context baseline, consistent with its lower error on rock pixels (Tab.~\ref{tab:rock_mae_co2}).
            }
            \label{fig:rollout_co2}
        \end{figure}

        We complement the aggregate error metrics with selected trajectories to illustrate how prediction errors develop spatially. In the FlowBench example~(Fig.~\ref{fig:rollout_flowbench}), both \texttt{PAdaNorm} and the parameter-matched wide In-context baseline initially reproduce the pressure pattern downstream of the obstacle. At later steps, we observe differences in the positions and shapes of these structures, together with distortions near the obstacle. In the acoustic example~(Fig.~\ref{fig:rollout_inclusions}), we observe discrepancies in the scattered wave pattern and grid-aligned artifacts as the rollout progresses. On CO\textsubscript{2}~(Fig.~\ref{fig:rollout_co2}), both models initially reproduce the main flow paths, but the depth-matched In-context baseline progressively predicts spurious velocity inside the rock, whereas \texttt{PAdaNorm} does so noticeably less. Neither model reproduces a large invasion region that develops later in the trajectory, illustrating how an early error in the CO\textsubscript{2} front propagates to all future rollout steps.

    \newpage
    \subsection{Methods}
        \label{app:methods}
        Tab.~\ref{tab:conditioning_acoustic} compares different approaches to conditioning on the one-way coupled fields; we describe each of them here.

        Baseline methods:
        \begin{enumerate}
            \item \textbf{In-context}: The conditioning fields are concatenated to the state as additional input channels before patch embedding, hence the transformer must infer the effect of the conditioning fields from context alone.
            \item \textbf{In-context (wide/deep)}: Parameter-matched models that raise the capacity of the In-context baseline. \textit{Wide} has an increased embedding dimension, \textit{deep} a larger number of blocks. $\dagger$ targets the parameter count of the spatial and temporal \texttt{PAdaNorm} variant. $\ddagger$ targets the spatial-only \texttt{PAdaNorm} variant.
            \item \textbf{AdaNorm (global)}: The conditioning embedding here is obtained from a single globally average-pooled conditioning vector, which a single MLP, shared across all blocks, maps to shift, scale, and gate parameters for both attention sublayers. This provides spatially uniform modulation: the same conditioning signal is broadcast to every token regardless of its spatial position.
            \item \textbf{GNOT}: see App.~\ref{subsubsec:gnot} for extensive details on our implementation of GNOT for the present setting.
        \end{enumerate}

        All \texttt{PAdaNorm} variants partition the conditioning fields into patches aligned with the state patches and produce a spatially varying conditioning vector per patch position. A shared MLP maps each conditioning patch to a conditioning embedding, which is then projected into the adaptive normalization modulation parameters (shift, scale, gate) that are applied to the normalization and residual connection along the specified attention axis. Tab.~\ref{tab:padaln_variants} describes each variant along the following components:
        \begin{itemize}
            \item \emph{Spatial modulation}: whether the spatial attention sublayer is modulated by \texttt{PAdaNorm}.
            \item \emph{Temporal modulation}: whether the temporal attention sublayer is modulated by \texttt{PAdaNorm}.
            \item \emph{Axis-specific}: temporal and spatial attention receive distinct modulation parameters via separate projections; otherwise, a single projection output is broadcast to both axes.
            \item \emph{Block-specific}: each transformer block owns its own projection; otherwise, a single MLP is shared across all blocks.
        \end{itemize}

        \begin{table}[ht!]
        \centering
        \caption{%
        \texttt{PAdaNorm} conditioning variants.
        }
        \label{tab:padaln_variants}
        \small
        \begin{tabular}{lcccc}
        \toprule
        Conditioning
        & Spatial modulation
        & Temporal modulation
        & Axis-specific
        & Block-specific \\
        \midrule
        Spatial-only            & \cmark     & \xmark     & {--}       & \cmark \\
        Temporal-only           & \xmark     & \cmark     & {--}       & \cmark \\
        Temporal (shared proj.) & \cmark     & \cmark     & \xmark     & \cmark \\
        Temporal + Spatial      & \cmark     & \cmark     & \cmark     & \cmark \\
        \midrule
        Fully shared            & \cmark     & \cmark     & \xmark     & \xmark \\
        Fully shared (spatial)  & \cmark     & \xmark     & {--}       & \xmark \\
        Fully shared (temporal) & \xmark     & \cmark     & {--}       & \xmark \\
        Shared per-axis         & \cmark     & \cmark     & \cmark     & \xmark \\
        \bottomrule
        \end{tabular}
        \end{table}

    \subsubsection{GNOT Baseline}
    \label{subsubsec:gnot}

        We re-implement the General Neural Operator Transformer (GNOT)~\citep{hao2023gnotgeneralneuraloperator} as a baseline, adapting it to our 2D~PDE surrogate setting. Below, we describe the key implementation choices and how they relate to the original paper.

        \paragraph{Tokenization}
        GNOT operates on a set of query points rather than on a regular pixel grid. To enable a fair comparison with our patch-based transformer, we tokenize the input field using the same \texttt{PatchEmbed}/\texttt{PatchUnembed} modules as our main model.  2D grid coordinates (normalized to $[0,1]$) are concatenated to each patch token and linearly projected back to the hidden dimension. The conditioning fields are embedded using the same \texttt{PatchEmbed} module as our main model, producing per-patch branch tokens for cross-attention conditioning.

        \paragraph{Temporal handling} Following common practice in neural-operator surrogates~\citep{li2021fourierneuraloperator}, we channel-concatenate the $T_{\mathrm{in}}$ input timesteps, reshaping $(B, T_{\mathrm{in}}, C, H, W)$ to
        $(B, T_{\mathrm{in}}\!\cdot\!C, H, W)$ before tokenization.  This avoids
        introducing a temporal attention axis absent from the original
        GNOT design.

        \paragraph{Conditioning on geometry}
        GNOT natively conditions on heterogeneous inputs through
        cross-attention with separate key/value projections per input
        type. We use this mechanism directly: the embedded conditioning
        fields are provided as a single-branch input to
        \texttt{LinearCrossAttention}.  The conditioning
        branch tokens share the same patch grid as the state tokens.

        \paragraph{Attention}
        We use GNOT's linear attention formulation, which applies softmax normalization along the \emph{feature} dimension of $Q$ and $K$ rather than the token dimension:
        \begin{equation}
            \mathrm{Attn}(Q, K, V)
            = \sigma_d(Q)\,\bigl[\sigma_d(K)^\top V\bigr],
        \end{equation}
        where $\sigma_d$ denotes row-wise softmax over the head
        dimension~$d$.  This yields $O(N d^2)$ complexity instead of the $O(N^2 d)$ of standard attention.

        \paragraph{Transformer block}
        Each GNOT block follows the structure:
        \begin{align}
          \mathbf{x} &\leftarrow \mathbf{x}
            + \mathrm{CrossAttn}\!\bigl(\mathrm{LN}(\mathbf{x}),\,
              \mathrm{LN}(\mathbf{y})\bigr), \\
          \mathbf{x} &\leftarrow \mathbf{x}
            + \mathrm{FFN}_1\!\bigl(\mathrm{LN}(\mathbf{x})\bigr), \\
          \mathbf{x} &\leftarrow \mathbf{x}
            + \mathrm{SelfAttn}\!\bigl(\mathrm{LN}(\mathbf{x})\bigr), \\
          \mathbf{x} &\leftarrow \mathbf{x}
            + \mathrm{FFN}_2\!\bigl(\mathrm{LN}(\mathbf{x})\bigr),
        \end{align}
        where each FFN is a residual MLP with GELU activations and $4\times$ expansion ratio.

        \paragraph{Mixture of Experts}
        The original GNOT supports geometric-gating MoE, where each FFN is replaced by $K$ expert MLPs gated by the spatial coordinates: $\mathrm{gate}(\mathbf{x}_{\mathrm{pos}})
        = \mathrm{softmax}\!\bigl(W \mathbf{x}_{\mathrm{pos}}\bigr)$. In our experiments, we use $K\!=\!1$, matching the non-MoE ablation in the original paper and found no consistent improvement when using MoE's.

        \paragraph{Positional encoding}
        GNOT supports optional Fourier positional features for the
        spatial coordinates. We implement this as a configurable option (\texttt{horiz\_fourier\_dim}; default~0, i.e.\ raw
        coordinates). In our experiments, we use raw coordinates, as
        preliminary tests showed no benefit from Fourier features on our datasets.

        \paragraph{Loss and evaluation}
        We train GNOT with the same loss as our main model on each benchmark---$L_1$ throughout, except the $L_2$ used on CO\textsubscript{2}---predicting a per-step delta in z-score space. Evaluation uses a 60-step autoregressive rollout with vRMSE as the primary metric, identical to all other baselines.

        \paragraph{Parameter budget}
        We size the GNOT architecture to approximately match the parameter count of our main model on each benchmark:
        \begin{itemize}
            \item Acoustic scattering ($256\!\times\!256$):
            $d\!=\!256$, $N_{\mathrm{blocks}}\!=\!5$, 1~MLP hidden layer $\approx$~18.8M
            parameters (target: $<$20M).
            \item FlowBench FPO ($256\!\times\!1024$):
            $d\!=\!256$, $N_{\mathrm{blocks}}\!=\!6$, 2~MLP hidden layers $\approx$~36M
            parameters (target: $\sim$39M).
            \item CO\textsubscript{2}--water flow in porous media ($512\!\times\!512$):
            $d\!=\!256$, $N_{\mathrm{blocks}}\!=\!6$, 2~MLP hidden layers $\approx$~35M
            parameters (target: $\sim$37.7M).
        \end{itemize}
        All configurations use a single attention head, no MoE, and no Fourier positional features.

        \paragraph{Training}
        On acoustic scattering and FlowBench, we train GNOT for 50k steps on a single GPU with
        a cosine learning-rate schedule (peak $1.875\!\times\!10^{-4}$,
        100-step warmup) and gradient clipping at norm~10.
        Validation is performed every 500 steps with 60-step rollouts. On CO\textsubscript{2}, GNOT instead uses the same optimization and evaluation settings as the other CO\textsubscript{2} models (Tab.~\ref{tab:hyperparameters_co2}): $T_{\mathrm{in}}{=}4$, a constant learning rate of $5{\times}10^{-5}$ after warmup, gradient clipping at norm~2.0, and validation every 1000 steps.

    \newpage
    \subsection{Experimental Settings}
    \label{app:experimental-settings}
        \paragraph{Compute}
            All experiments were run on NVIDIA H100 GPUs.
            Number of GPUs used for training and approximate wall-clock time per run were: i) Acoustic scattering: 1 GPU, 9--10 hours; ii) FlowBench: 1 GPU, 8--10 hours; iii) CO\textsubscript{2}: 4 GPUs, 13--20 hours (2 hours for GNOT).
        \begin{table}[ht!]
\centering
\caption{%
  Configuration for the acoustic scattering experiments from
  \href{https://polymathic-ai.org/the_well/}{The Well}~\citep{thewelldataset2024}
  in Table~\ref{tab:main_aggregated}.
  All runs share the settings in the upper block and differ only in the lower one;
  head dimension is held at 128 throughout, so width and head count move together.
  Normalization uses The Well's \texttt{ZScoreNormalization} with statistics computed
  on the training split.
  GNOT baselines use separate architectures and training settings, given in App.~\ref{subsubsec:gnot}.
}
\label{tab:hyperparameters_acoustic}
\small
\setlength{\tabcolsep}{5pt}
\begin{tabular}{l l}
\toprule
\multicolumn{2}{l}{\emph{Data}} \\
Datasets & acoustic scattering: maze, inclusions, discontinuous ($256{\times}256$) \\
Split & official The Well \texttt{train} / \texttt{valid} splits \\
Joint training & the three datasets concatenated, sampled in proportion to size \\
State fields & pressure and velocity (3 channels), z-score normalized \\
Conditioning channels & density and speed of sound (2 channels) \\
Excluded trajectories & one trajectory (index 347) in inclusions, whose simulation diverges \\
\midrule
\multicolumn{2}{l}{\emph{Architecture}} \\
Patch size & 8 ($32{\times}32 = 1024$ patches) \\
Input timesteps $T_{\mathrm{in}}$ & 3 \\
Head dimension & 128 \\
MLP ratio & 4.0 \\
QK-norm & enabled \\
Temporal RoPE base $\theta$ & 10\,000 \\
Spatial RoPE base $\theta$ & 10\,000 \\
Cross-attention RoPE base $\theta$ & 300 (cross-attention row only; see below) \\
\midrule
\multicolumn{2}{l}{\emph{Optimization}} \\
Optimizer & AdamW, $\beta{=}(0.9, 0.999)$, $\varepsilon{=}10^{-8}$, weight decay 0.01 \\
Learning rate & $1.875{\times}10^{-4}$, constant after warmup \\
Warmup & 100 steps, linear from $10^{-7}$ \\
Training steps & 50\,000 \\
Batch size & 32 \\
Gradient clipping & global norm 10.0 \\
Loss & per-field $L_1$ \\
Precision & fp32 \\
Training seeds & 1234, 21, 67 \\
\midrule
\multicolumn{2}{l}{\emph{Evaluation}} \\
Validation windows & 200 per dataset (single-dataset runs: 200; joint: 600) \\
Window selection & uniformly at random from all validation windows (seed 42); identical across runs \\
Joint aggregation & mean over the three datasets, each contributing 200 windows \\
\bottomrule
\end{tabular}

\vspace{0.7em}

\begin{tabular}{l r r r l}
\toprule
Model & $d_{\mathrm{model}}$ & Blocks & Heads & Conditioning module \\
\midrule
In-context & 256 & 10 & 2 & {--} \\
In-context$^\ddagger$ (wide) & 384 & 6 & 3 & {--} \\
In-context$^\ddagger$ (deep) & 256 & 14 & 2 & {--} \\
AdaNorm (global) & 256 & 10 & 2 & MLP $256{\to}512$, 4 layers, global pool \\
Cross-attention & 256 & 11 & 2 & MLP $256{\to}512$, 4 layers, per-patch KV \\
\texttt{PAdaNorm} (spatial) & 256 & 10 & 2 & MLP $256{\to}512$, 4 layers, per-patch \\
\bottomrule
\end{tabular}
\end{table}

        \begin{table}[ht!]
\centering
\caption{%
  Configuration for the
  \href{https://github.com/baskargroup/FlowBench}{FlowBench FPO}~\citep{tali2024flowbench}
  experiments in Table~\ref{tab:main_aggregated}.
  All runs share the settings in the upper block and differ only in the lower one;
  conditioning inputs are normalized with statistics computed on the training split.
  GNOT baselines use separate architectures and training settings, given in App.~\ref{subsubsec:gnot}.
}
\label{tab:hyperparameters_flowbench}
\small
\setlength{\tabcolsep}{5pt}
\begin{tabular}{l l}
\toprule
\multicolumn{2}{l}{\emph{Data}} \\
Dataset & FlowBench FPO, 2D Navier--Stokes, $1024{\times}256$ \\
Split (simulations) & 881 train / 219 valid, stratified by geometry type \\
Trajectory length & 242 steps; 236 sliding windows per trajectory \\
State fields & $u, v, p$ (3 channels), z-score normalized\\
Conditioning channels & mask, SDF, Reynolds number (broadcast as a uniform 2D field) \\
Reynolds preprocessing & $\log_{10}$ then z-score normalized \\
SDF preprocessing & z-score normalized \\
Mask preprocessing & none (already binary) \\
\midrule
\multicolumn{2}{l}{\emph{Architecture }} \\
Patch size & 16 ($64{\times}16 = 1024$ patches) \\
Input timesteps $T_{\mathrm{in}}$ & 6 \\
Head dimension & 128 \\
MLP ratio & 4.0 \\
QK-norm & enabled \\
Temporal RoPE base $\theta$ & 30 \\
Spatial RoPE base $\theta$ & 10\,000 \\
Cross-attention RoPE base $\theta$ & 60 (cross-attention row only; see below) \\
\midrule
\multicolumn{2}{l}{\emph{Optimization}} \\
Optimizer & AdamW, $\beta{=}(0.9, 0.999)$, $\varepsilon{=}10^{-8}$, weight decay 0.01, fused \\
Learning rate & $4.6875{\times}10^{-5}$, constant after warmup \\
Warmup & 100 steps, linear from $10^{-7}$ \\
Training steps & 50\,000 \\
Batch size & 8 \\
Gradient clipping & global norm 10.0 \\
Loss & per-field $L_1$ \\
Precision & fp32 \\
Training seeds & 1234, 21, 67 \\
\midrule
\multicolumn{2}{l}{\emph{Evaluation}} \\
Validation windows & 60 \\
Window selection & uniformly at random from all validation windows (seed 42); identical across runs \\
\bottomrule
\end{tabular}

\vspace{0.7em}

\begin{tabular}{l r r r l}
\toprule
Model & $d_{\mathrm{model}}$ & Blocks & Heads & Conditioning module \\
\midrule
In-context & 384 & 10 & 3 & {--} \\
In-context$^\ddagger$ (wide) & 512 & 7 & 4 & {--} \\
In-context$^\ddagger$ (deep) & 384 & 13 & 3 & {--} \\
AdaNorm (global) & 384 & 10 & 3 & MLP $256{\to}512$, 4 layers, global pool \\
Cross-attention & 384 & 11 & 3 & MLP $256{\to}512$, 4 layers, per-patch KV \\
\texttt{PAdaNorm} (spatial) & 384 & 10 & 3 & MLP $256{\to}512$, 4 layers, per-patch \\
\bottomrule
\end{tabular}
\end{table}

        \begin{table}[ht!]
\centering
\caption{%
  Configuration for the \href{https://doi.org/10.5061/dryad.jm63xsjn5}{CO\textsubscript{2}--water flow in porous media}~\citep{abdellatif2025co2pore} experiments in
  Table~\ref{tab:main_aggregated}.
  All runs share the settings in the upper block and differ only in the lower one;
  conditioning inputs are normalized with statistics computed on the training split.
  GNOT baselines use separate architectures and training settings, given in App.~\ref{subsubsec:gnot}.
}
\label{tab:hyperparameters_co2}
\small
\setlength{\tabcolsep}{5pt}
\begin{tabular}{l l}
\toprule
\multicolumn{2}{l}{\emph{Data}} \\
Dataset & CO\textsubscript{2}--water flow in porous media, $512{\times}512$ \\
Split (simulations) & 500 train / 124 valid, split by base geometry \\
Split procedure & 100 base geometries permuted with seed 42; 80 train / 20 valid \\
Trajectory length & 100 steps; 96 sliding windows per trajectory \\
State fields & $\alpha_{\mathrm{water}}, p, u_x, u_y$ (4 channels), z-score normalized \\
Conditioning channels & binary pore mask (1 = pore, 0 = rock) and normalized $x$, $y$ coordinates in $[-1, 1]$ (3 channels) \\
Mask preprocessing & none (already binary) \\
\midrule
\multicolumn{2}{l}{\emph{Architecture}} \\
Patch size & 8 ($64{\times}64 = 4096$ patches) \\
Input timesteps $T_{\mathrm{in}}$ & 4 \\
Head dimension & 128 \\
MLP ratio & 4.0 \\
QK-norm & enabled \\
Temporal RoPE base $\theta$ & 9 \\
Spatial RoPE base $\theta$ & 60 \\
Cross-attention RoPE base $\theta$ & 60 (cross-attention row only) \\
\midrule
\multicolumn{2}{l}{\emph{Optimization}} \\
Optimizer & AdamW, $\beta{=}(0.9, 0.999)$, $\varepsilon{=}10^{-8}$, weight decay 0.01, fused \\
Learning rate & $5{\times}10^{-5}$, constant after warmup \\
Warmup & 100 steps, linear from $10^{-7}$ \\
Training steps & 50\,000 \\
Batch size & 16 (4 GPUs $\times$ 4) \\
Gradient clipping & global norm 2.0 \\
Loss & $L_2$ over all pixels (not masked to the pore space) \\
Precision & fp32 \\
Training seeds & 1234, 21, 67 \\
\midrule
\multicolumn{2}{l}{\emph{Evaluation}} \\
Validation windows & 60 \\
Window selection & uniformly at random from all validation windows (seed 42); identical across runs \\
Rollout horizon & 60 steps, predicted deltas \emph{not} masked to the pore space \\
Validation interval & 1000 steps (checkpointing tied to the same grid) \\
Checkpoint selection & best unmasked $T{\in}[21{:}60]$ vRMSE over the 50 checkpoints, per seed \\
\bottomrule
\end{tabular}

\vspace{0.7em}

\begin{tabular}{l r r r l}
\toprule
Model & $d_{\mathrm{model}}$ & Blocks & Heads & Conditioning module \\
\midrule
In-context & 384 & 10 & 3 & {--} \\
In-context$^\ddagger$ (wide) & 512 & 7 & 4 & {--} \\
In-context$^\ddagger$ (deep) & 384 & 13 & 3 & {--} \\
AdaNorm (global) & 384 & 10 & 3 & MLP $256{\to}512$, 4 layers, global pool \\
Cross-attention & 384 & 10 & 3 & MLP $256{\to}512$, 4 layers, per-patch KV \\
\texttt{PAdaNorm} (spatial) & 384 & 10 & 3 & MLP $256{\to}512$, 4 layers, per-patch \\
\bottomrule
\end{tabular}
\end{table}

    \newpage
    \subsection{Computational Cost of Conditioning}
    \label{app:flops}

        This section derives the forward cost of each strategy and reports measured counts for the exact configurations that were trained (Tab.~\ref{tab:flops}).

        \paragraph{Counting convention}
            For each architecture in the paper, we count forward-pass FLOPs with PyTorch's \texttt{FlopCounterMode}. These only account for matrix multiplications, convolutions, and attention matrix products; the element-wise operations we do not count (normalization, activations, and the modulation itself) amount to less than $1\%$ of the total in our measurements.

            We write $N = N_xN_y$ for the number of patches per frame, $T_{\mathrm{in}}$ for the history length, $D$ for the model width, $D_{\mathrm{ff}} = rD$ for the feed-forward width ($r = 4$), $N_{\mathrm{blocks}}$ for the number of blocks, $D_h$ for the width of the conditioning encoder, $L_c$ for the number of its hidden layers, $D_c$ for the embedding dimension of a conditioning token, and $C_{\mathrm{state}}, C_{\mathrm{cond}}$ for the number of state and conditioning channels. A linear map $\mathbb{R}^{a} \to \mathbb{R}^{b}$ applied to $M$ tokens costs $2Mab$, and attention between $M$ queries and $M'$ keys of total width $D$ costs $4MM'D$, counting $QK^\top$ and the value matrix product.

        \paragraph{Backbone}
            Here, we derive the FLOPs for the forward pass of the model backbone, i.e., ignoring the conditioning mechanisms.
            Temporal attention reshapes the tokens to $(N, T_{\mathrm{in}}, D)$ and applies four $D \to D$ projections followed by attention over $T_{\mathrm{in}}$ positions, which costs $8T_{\mathrm{in}}ND^2 + 4NT_{\mathrm{in}}^2D$ per block. The spatial sublayer of Eq.~\ref{eq:spatial_block} uses a single fused projection $D \to 3D + 2D_{\mathrm{ff}}$ shared by the attention and feed-forward branches, two output projections, and attention over $N$ positions, costing $T_{\mathrm{in}}ND(8D + 6D_{\mathrm{ff}}) + 4T_{\mathrm{in}}N^2D$. With $r = 4$ the two sublayers together give
            \begin{equation}
                \Phi_{\mathrm{block}}
                = \underbrace{40\,T_{\mathrm{in}}ND^2}_{\text{projections}}
                + \underbrace{4\,T_{\mathrm{in}}N^2D}_{\text{spatial attention}}
                + \underbrace{4\,NT_{\mathrm{in}}^2D}_{\text{temporal attention}},
                \label{eq:flops_block}
            \end{equation}
            Outside the transformer blocks, the tokenizer costs $2T_{\mathrm{in}}NDCP^2$ with $C = C_{\mathrm{state}} + C_{\mathrm{cond}}$ for In-context and $C = C_{\mathrm{state}}$ otherwise, and the decoder, applied to the last frame only, costs $2ND\frac{D}{4}s_1^2 + 2Ns_1^2\frac{D}{4}C_{\mathrm{state}}s_2^2$ for the two transposed convolutions of stride $s_1s_2 = P$. Every model of Tab.~\ref{tab:main_aggregated} except GNOT shares this architecture; they differ only in the conditioning term below. GNOT's FLOPs are measured directly (Tab.~\ref{tab:flops}).

        \paragraph{Global AdaNorm}
            Average pooling reduces the conditioning fields to a single vector before the encoder, so the whole conditioning path acts on one token and one shared modulation network:
            \begin{equation}
                \Phi_{\mathrm{AdaNorm}} = 2\big(C_{\mathrm{cond}}D_h + (L_c-1)D_h^2 + D_hD_c\big) + 2\big(D_cD_h + (L_c-1)D_h^2 + 3D_hD\big),
            \end{equation}
            which is $O(1)$ in both $N$ and $T_{\mathrm{in}}$, and hence only $0.004$ GFLOPs.

        \paragraph{\texttt{PAdaNorm}}
            The shared encoder $f_\theta$ maps each of the $N$ patches from $C_{\mathrm{cond}}P^2$ raw pixels to $D_h$, and each block applies its own linear projection $g^{(\ell)}_\phi: D_h \to 3D$ per patch:
            \begin{equation}
                \Phi_{\texttt{PAdaNorm}}
                = \underbrace{2N\big(C_{\mathrm{cond}}P^2D_h + 2D_cD_h + 2(L_c-1)D_h^2\big)}_{f_\theta,\;\text{once per forward}}
                + \underbrace{6\,N_{\mathrm{blocks}}ND_hD}_{g^{(\ell)}_\phi,\;\text{one per block}}\,.
                \label{eq:flops_padaln}
            \end{equation}
            The resulting cost is \emph{linear} in the number of patches, \emph{independent of the history length} because one set of modulation parameters is broadcast over all $T_{\mathrm{in}}$ timesteps, and is also independent of the rollout length when the conditioning is static.

        \paragraph{Cross-attention}
            The conditioning fields are tokenized on the same patch grid by a dedicated conditioning encoder, and each block allows the $T_{\mathrm{in}}N$ state tokens to attend to the conditioning tokens via query, key, value, and output projections. Hence
            \begin{equation}
                \Phi_{\mathrm{cross}}
                = 2N\big(C_{\mathrm{cond}}P^2D_h + (L_c-1)D_h^2 + D_hD_c\big)
                + 4\,N_{\mathrm{blocks}}T_{\mathrm{in}}ND\,(D + D_c + N),
                \label{eq:flops_cross}
            \end{equation}
            where the three terms in the bracket are the query and output projections, the key and value projections, and the attention products. Unlike Eq.~\ref{eq:flops_padaln}, the computation cost of cross-attention is linear in $T_{\mathrm{in}}$ and quadratic in $N$, and in cases where conditioning is static, only the key and value projections are independent of the rollout length.

        \paragraph{Scaling}
            Ignoring the encoder, the ratio of the per-block conditioning costs of cross-attention Eq.~\ref{eq:flops_cross} and \texttt{PAdaNorm} Eq.~\ref{eq:flops_padaln} is significantly in favor of \texttt{PAdaNorm}
            \begin{equation}
                \frac{4T_{\mathrm{in}}ND(D + D_c + N)}{6ND_hD} = \frac{2T_{\mathrm{in}}(D + D_c + N)}{3D_h},
            \end{equation}
            equal to $6.0$ on the acoustic configuration, $13.0$ on FlowBench, where the history is twice as long, and $24.7$ on CO\textsubscript{2}, which has four times as many patches per frame and four input timesteps.

        \paragraph{Amortization over a rollout}
            The conditioning fields are static for the acoustic scattering, FlowBench, and CO\textsubscript{2} datasets, allowing some of the conditioning computations to be reused across the rollout. For \texttt{PAdaNorm}, the entire conditioning computation can be cached once and reused: the computation cost of an $S$-step rollout is $S\,\Phi_{\mathrm{backbone}} + \Phi_{\texttt{PAdaNorm}}$, so its marginal per-step cost in counted FLOPs equals that of an unconditioned model of the same size; the only remaining per-step overhead is the element-wise modulation, which \texttt{FlopCounterMode} does not count. For cross-attention, only the key and value projections can be reused across rollout steps; queries, attention products, and the output projection depend on the current state and are done $S$ times. The gap, therefore, grows linearly with the rollout length: cross-attention repeats its state-dependent conditioning cost at every step, whereas \texttt{PAdaNorm} pays its conditioning cost once (Tab.~\ref{tab:flops}).
        \begin{table}[t!]
\centering
\caption{%
  Forward-pass and rollout cost of each conditioning strategy, for the
  models reported in Tab.~\ref{tab:main_aggregated}.
  Batch size~1, one multiply--add counted as two FLOPs.
  \emph{cond.\ (static)} is the part of the conditioning cost that does not depend on the
  state and is computed once per rollout; \emph{cond.\ (state-dep.)} is the part recomputed at every
  step, such as the cross-attention queries. The rollout columns are accordingly $\Phi_{\mathrm{step}} = \Phi_{\mathrm{fwd}} - \Phi_{\mathrm{static}}$ and
  $60\,\Phi_{\mathrm{step}} + \Phi_{\mathrm{static}}$.
  App.~\ref{app:flops} gives the counting convention and the derivations.
}
\label{tab:flops}
\small
\setlength{\tabcolsep}{3pt}
\oursbox{ours-flops-acoustic}
\oursbox{ours-flops-flowbench}
\oursbox{ours-flops-co2}
\begin{tabular}{@{}l r r r r r r@{}}
\toprule
\multirow{2}{*}{Conditioning}
  & \multirow{2}{*}{\#Params}
  & \multicolumn{3}{c}{Forward pass (GFLOPs)}
  & \multicolumn{2}{c}{60-step rollout}
  \\
\cmidrule(lr){3-5}\cmidrule(lr){6-7}
  & & total
  & \shortstack{cond.\\(static)} & \shortstack{cond.\\(state-dep.)}
  & \shortstack{GFLOPs\\per step} & TFLOPs \\
\midrule
\multicolumn{7}{l}{\emph{Acoustic scattering --- $256{\times}256$, $P{=}8$, $T_{\mathrm{in}}{=}3$}} \\
In-context                  & 13.3\,M & 113.5 & {--}   & {--}   & 113.5 & 6.81 \\
In-context (wide)           & 18.0\,M & 138.9 & {--}   & {--}   & 138.9 & 8.33 \\
In-context (deep)           & 18.6\,M & 158.6 & {--}   & {--}   & 158.6 & 9.52 \\
AdaNorm (global)            & 15.5\,M & 113.3 & $<0.01$ & {--}  & 113.3 & 6.80 \\
Cross-attention             & 18.5\,M & 179.7 & 10.9   & 44.3   & 168.9 & 10.14 \\
GNOT                        & 18.8\,M & 41.2  & 2.3    & 2.0    & 38.8  & 2.33 \\
\oursL{ours-flops-acoustic}\texttt{PAdaNorm} (spatial-only, ours) & 19.1\,M & 125.2 & 11.9 & {--} & 113.3 & 6.81\oursR{ours-flops-acoustic}\oursB{ours-flops-acoustic} \\
\midrule
\multicolumn{7}{l}{\emph{FlowBench FPO --- $1024{\times}256$, $P{=}16$, $T_{\mathrm{in}}{=}6$}} \\
In-context                  & 30.8\,M & 468.2 & {--}   & {--}   & 468.2 & 28.09 \\
In-context (wide)           & 38.6\,M & 553.7 & {--}   & {--}   & 553.7 & 33.22 \\
In-context (deep)           & 39.6\,M & 606.1 & {--}   & {--}   & 606.1 & 36.36 \\
AdaNorm (global)            & 32.9\,M & 464.6 & $<0.01$ & {--}  & 464.6 & 27.87 \\
Cross-attention             & 40.2\,M & 686.0 & 29.3   & 146.2  & 656.7 & 39.43 \\
GNOT                       & 36.1\,M & 77.1  & 3.1    & 2.4    & 74.1  & 4.45 \\
\oursL{ours-flops-flowbench}\texttt{PAdaNorm} (spatial-only, ours) & 38.6\,M & 481.2 & 16.6 & {--} & 464.6 & 27.89\oursR{ours-flops-flowbench}\oursB{ours-flops-flowbench} \\
\midrule
\multicolumn{7}{l}{\emph{CO\textsubscript{2}--water flow in porous media --- $512{\times}512$, $P{=}8$, $T_{\mathrm{in}}{=}4$}} \\
In-context                  & 29.9\,M & 2005.2 & {--}   & {--}   & 2005.2 & 120.31 \\
In-context (wide)           & 37.3\,M & 2175.5 & {--}   & {--}   & 2175.5 & 130.53 \\
In-context (deep)           & 38.8\,M & 2604.7 & {--}   & {--}   & 2604.7 & 156.28 \\
AdaNorm (global)            & 32.3\,M & 2002.8 & $<0.01$ & {--}  & 2002.8 & 120.17 \\
Cross-attention             & 35.8\,M & 3203.0 & 72.7   & 1127.4 & 3130.2 & 187.89 \\
GNOT                        & 35.3\,M & 301.2  & 11.1   & 9.7    & 290.0  & 17.41 \\
\oursL{ours-flops-co2}\texttt{PAdaNorm} (spatial-only, ours) & 37.7\,M & 2067.0 & 64.2 & {--} & 2002.8 & 120.23\oursR{ours-flops-co2}\oursB{ours-flops-co2} \\
\bottomrule
\end{tabular}
\end{table}

\newpage

\section{Ablation details}
\label{app:ablation}

    \subsection{Global AdaNorm Baseline}
    \label{app:adaln_baseline}

        As one of the baselines, we consider a globally conditioned adaptive-normalization strategy in which the spatially varying field $\mathbf{C} \in
        \mathbb{R}^{C_{\mathrm{cond}} \times H \times W}$ is first reduced to a single global conditioning vector via spatial average pooling:
        \begin{equation}
        \mathbf{c} = \frac{1}{HW} \sum_{p,q} \mathbf{C}_{:,p,q} \in \mathbb{R}^{C_{\mathrm{cond}}},
        \end{equation}
        followed by a shallow MLP $f_\theta$ that maps $\mathbf{c}$ to a latent representation $\mathbf{c}' = f_\theta(\mathbf{c}) \in
        \mathbb{R}^{D_c}$. Unlike \texttt{PAdaNorm}, this conditioning vector is spatially uniform: all tokens within a given sample receive
        identical modulation parameters, discarding the spatial structure of the conditioning fields.

        A single MLP $g_\phi$, shared across all blocks, then maps $\mathbf{c}'$ to scale, shift and gate parameters,
        \begin{equation}
        \boldsymbol{\gamma},\; \boldsymbol{\beta},\; \boldsymbol{\alpha} = g_\phi(\mathbf{c}'),
        \end{equation}
        which modulate both the temporal and the spatial attention sublayers of every block in the AdaLN-Zero fashion of Sec.~\ref{subsec:adaln}. The modulation is therefore identical across tokens, blocks, and attention axes.

    \subsection{Different approaches to patching}
    \label{app:different_patching}

        \paragraph{Sliding-window MLP}
            We keep a unique conditioning patch for each state-variable patch, with the conditioning patch overlapping neighboring patches to include more information.
            Let $\mathbf{C} \in \mathbb{R}^{C_{\mathrm{cond}} \times H \times W}$ denote the stacked conditioning fields, $P$ the patch size, and $k > P$ the kernel size. For each patch index $(i, j)$ on the grid of size $N_x \times N_y$, we extract a sliding window of size $k \times k$, centered on the corresponding patch and zero-padded at the boundaries. A shared MLP then maps each flattened window to the conditioning space: $\mathbf{c}_{ij} = \mathrm{MLP}(\mathbf{w}_{ij}) \in \mathbb{R}^{D_h}$. The receptive field is exactly $k \times k$ pixels per token. Because the MLP operates on the full flattened window, it can learn position-specific patterns within each receptive field, at the cost of a first-layer parameter count that grows as $\mathcal{O}(C_{\mathrm{cond}}\, k^2\, D_h)$.

        \paragraph{Convolutional encoder}
            We stack $L_{\mathrm{conv}}$ convolutional layers (kernel size $k$, same-padding, SiLU activations) at full spatial resolution, followed by a single strided convolution (kernel size $P$, stride $P$) that downsamples to the patch grid. The effective receptive field is $P + L_{\mathrm{conv}}(k - 1)$ pixels per token and grows linearly with depth, while weight sharing keeps the per-layer cost at $\mathcal{O}(D_h\, k^2)$, independent of the receptive field size. The trade-off is translation equivariance: unlike the MLP variant, the convolutional encoder cannot learn position-specific patterns within each patch.

        \paragraph{Coarse conditioning patches}
              While the former two approaches produce a unique conditioning vector per state patch, we also consider reducing the conditioning resolution relative to the state grid. Given an integer multiplier $m \geq 1$, we partition the conditioning fields into non-overlapping patches of size $mP \times mP$, each covering $m \times m$ state patches. A shared MLP maps each coarse patch to a conditioning vector $\mathbf{c}_s \in \mathbb{R}^{D_h}$, which is then broadcast identically to all $m^2$ state tokens it covers. Note that $m = 1$ recovers the default one-to-one alignment. Unlike the two overlapping variants, the coarse patches are not centered on the state patch they condition on, and we require $m$ to divide both $N_x$ and $N_y$.

        \begin{figure}[ht]
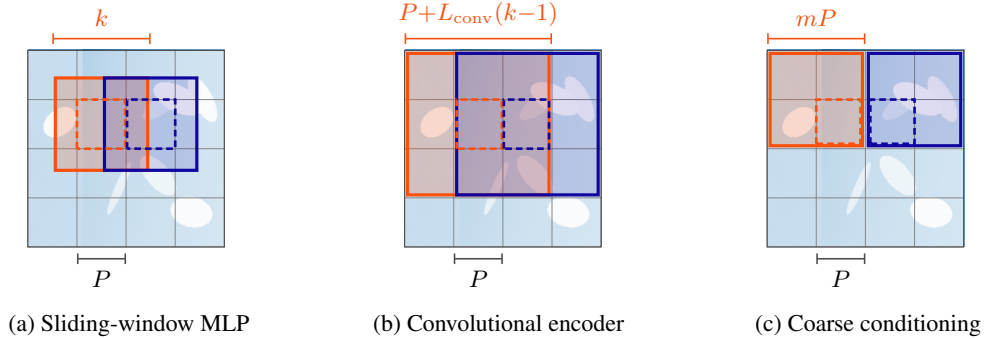

            \centering
            \begin{subfigure}[b]{0.3\textwidth}
                \centering
                \input{figures/sliding_window}
                \caption{Sliding-window MLP}
            \end{subfigure}
            \hfill
            \begin{subfigure}[b]{0.3\textwidth}
                \centering
                \input{figures/conv}
                \caption{Convolutional encoder}
            \end{subfigure}
            \hfill
            \begin{subfigure}[b]{0.3\textwidth}
                \centering
                \input{figures/coarser_patch}
                \caption{Coarse conditioning}
            \end{subfigure}
            \caption{\textbf{Conditioning patching strategies.} Three approaches for conditioning patch design. Orange and blue regions show the input area contributing to two adjacent conditioning tokens, dashed squares mark the corresponding state patches, and the grid shows the $P \times P$ state patches. The width of the orange region is $k$ for the sliding window, $P + L_{\mathrm{conv}}(k-1)$ for the convolutional encoder, and $mP$ for coarse conditioning.}
            \label{fig:conditioning_patches}
      \end{figure}

\end{document}